\PassOptionsToPackage{table}{xcolor}
\documentclass[runningheads]{llncs}

\usepackage[preprint]{colm} 

\setcitestyle{authoryear}

\usepackage{graphicx}
\usepackage{booktabs}
\usepackage{adjustbox}
\usepackage{threeparttable}
\usepackage[accsupp]{axessibility}
\usepackage{xltabular} 
\usepackage{microtype}
\usepackage{tabularx}
\usepackage{amssymb}
\usepackage{paralist}
\usepackage{todonotes}
\usepackage{breakurl}
\usepackage{subcaption}
\usepackage{latexsym}
\usepackage{amsmath}
\usepackage{float}
\usepackage{footnote}
\usepackage{enumitem}
\usepackage{bm}
\usepackage{arydshln}
\usepackage{multicol}
\usepackage{multirow}
\usepackage{colortbl}
\usepackage{bbding}
\usepackage{makecell}
\usepackage{mathtools}
\usepackage{imakeidx}
\usepackage{longtable}
\usepackage{tabularx}
\usepackage{wrapfig}
\usepackage{rotating}
\usepackage[edges]{forest}
\usepackage[normalem]{ulem}
\usepackage{CJKutf8}
\usepackage{awesomebox}
\usepackage{diagbox}
\usepackage[most]{tcolorbox}

\RequirePackage{xspace}
\makeatletter
\DeclareRobustCommand\onedot{\futurelet\@let@token\@onedot}
\def\@onedot{\ifx\@let@token.\else.\null\fi\xspace}

\makeatother

\usepackage{tocloft}

\cftsetindents{section}{0em}{1.5em}
\cftsetindents{subsection}{1.5em}{2.3em}
\cftsetindents{subsubsection}{3.8em}{3.2em}


\usepackage{pifont}
\usepackage{lscape}
\usepackage{algorithm}
\usepackage{algpseudocode}

\usepackage{fancyhdr}
\renewcommand{\headrulewidth}{1pt}
\usepackage{fancyvrb}
\usepackage{fvextra}
\usepackage{amsfonts}
\usepackage{wrapfig}
\usepackage{setspace}
\usepackage{hyperref}
\usepackage{orcidlink}

\definecolor{mydarkblue}{rgb}{0,0.08,0.45}
\definecolor{wkblue}{rgb}{0.2, 0.3, 0.6}
\definecolor{meta-color}{rgb}{0.5, 0.5, 0.5}
\definecolor{darkblue}{rgb}{0, 0, 0.5}
\definecolor{geovistagray}{gray}{0.95}
\definecolor{myblue}{rgb}{0.9, 0.1, 0.94}
\definecolor{mygreen}{rgb}{0.64, 0.56, 0.88}
\definecolor{myyellow}{rgb}{0.68, 0.6, 0.1}
\definecolor{fancygreen}{rgb}{0.33, 0.68, 0.20}
\definecolor{salmon}{rgb}{0.94, 0.52, 0.49}
\definecolor{tablegreen}{rgb}{0.82, 0.94, 0.75}
\definecolor{tableblue}{rgb}{0.81, 0.90, 0.94}
\definecolor{tablered}{rgb}{0.97, 0.85, 0.85}
\definecolor{tableorange}{rgb}{0.96, 0.85, 0.81}
\definecolor{bestcolor}{RGB}{210, 222, 239}
\definecolor{secondcolor}{RGB}{234, 239, 247}
\definecolor{thirdcolor}{RGB}{193, 214, 229}
\definecolor{line-blue}{RGB}{243, 248, 252}
\definecolor{line-green}{RGB}{200,242,200}
\definecolor{line-red}{RGB}{255,215,215}
\definecolor{line-gray}{RGB}{242, 242, 242}
\definecolor{sensepurple}{HTML}{5D2DD6}
\definecolor{rynn}{RGB}{108,92,186} 

\newenvironment{itemize*}%
 {\leftmargini=10pt\begin{itemize}%
  \setlength{\itemsep}{0pt}%
  \setlength{\parskip}{0pt}%
  }%
 {\end{itemize}}
\newenvironment{enumerate*}%
 {\begin{enumerate}%
  \setlength{\itemsep}{0pt}%
  \setlength{\parskip}{0pt}}%
 {\end{enumerate}}

\hypersetup{
    hidelinks
}

\newcommand{\paragrapha}[2][3pt]{\vspace{#1}\noindent\textbf{#2}}

\newcolumntype{x}[1]{>{\centering\arraybackslashå}p{#1pt}}
\newlength\savewidth

\newcommand{\PreserveBackslash}[1]{\let\temp=\\#1\let\\=\temp}
\newcolumntype{C}[1]{>{\PreserveBackslash\centering}p{#1}}
\newcolumntype{L}[1]{>{\PreserveBackslash\raggedright}p{#1}}

\usepackage[dvipsnames]{xcolor} 
\usepackage{hyperref}
\hypersetup{
    colorlinks=true,
    linkcolor=RoyalBlue,
    citecolor=RoyalBlue,
    urlcolor=MidnightBlue
}

\definecolor{boxbackground}{HTML}{F3F4F7}  

\newtcolorbox{abstractbox}{
    colback=boxbackground,  
    colframe=boxbackground, 
    boxrule=0pt,            
    arc=4mm,                
    auto outer arc,
    left=18pt, right=18pt, top=18pt, bottom=18pt,            
    width=\linewidth,       
    halign=justify          
}

\begin{document}

\title{Hy-Embodied-VLM-1.0: Efficient Physical-World Agents}

\titlerunning{HY-Embodied}

\author{\textbf{Tencent Robotics X}~~~~~~\textbf{Hy Vision Team}~~~~~~\textbf{Futian Laboratory}}


\maketitle

\makeatletter
\def\@makefnmark{}
\makeatother


\vspace{-20pt}
\begin{abstractbox}

Building capable embodied agents requires not only multimodal perception and understanding, but also agentic capabilities for reasoning about actions, adapting to evolving situations, and interacting with the physical world. In this report, we introduce Hy-Embodied-VLM-1.0, an efficient and powerful embodied foundation model specifically designed for embodied agents operating in the physical world.
To cultivate such capabilities from the pre-training stage onward, we define an action-centric capability taxonomy comprising three progressive dimensions: Action-Relevant State Understanding, Action–Transition Reasoning, and Sequential and Adaptive Reasoning. Guided by this taxonomy, we develop a systematic data pipeline and curate data mixtures spanning both pre-training and post-training.
To deliver strong physical-world understanding and interaction capabilities while supporting latency-sensitive deployment, we build our model on the Hy3-A3B language backbone and the Hy-ViT2 vision encoder. Its efficient Mixture-of-Experts architecture combines strong model capacity with high inference efficiency. We evaluate Hy-Embodied-VLM-1.0 on a comprehensive suite of 38 benchmarks covering embodied perception, physical-world understanding, and embodied reasoning. The model achieves the best performance among similarly sized models on 19 of the 38 benchmarks and substantially outperforms strong competitors, including Qwen3.6-A3B and Cosmos 3. Compared with the previous-generation Hy-Embodied-0.5 MoT-2B, Hy-Embodied-VLM-1.0 improves average performance by 8.4\%. Despite activating only 3B parameters, it achieves performance close to that of the previous-generation model with 32B activated parameters. Beyond static benchmark evaluation, Hy-Embodied-VLM-1.0 also demonstrates strong performance on embodied agentic tasks requiring multi-turn interaction and long-horizon reasoning.

\vspace{1em} 
\textbf{Github:} \href{https://github.com/Tencent-Hunyuan/HY-Embodied}{{\small\nolinkurl{github.com/Tencent-Hunyuan/HY-Embodied}}}\\
\textbf{Model:} \href{https://huggingface.co/tencent/Hy-Embodied-VLM-1.0}{{\small\nolinkurl{huggingface.co/tencent/Hy-Embodied-VLM-1.0}}}
\end{abstractbox}

\vspace{-10pt}
\begin{figure}[htbp]
  \centering
  \includegraphics[width=\linewidth]{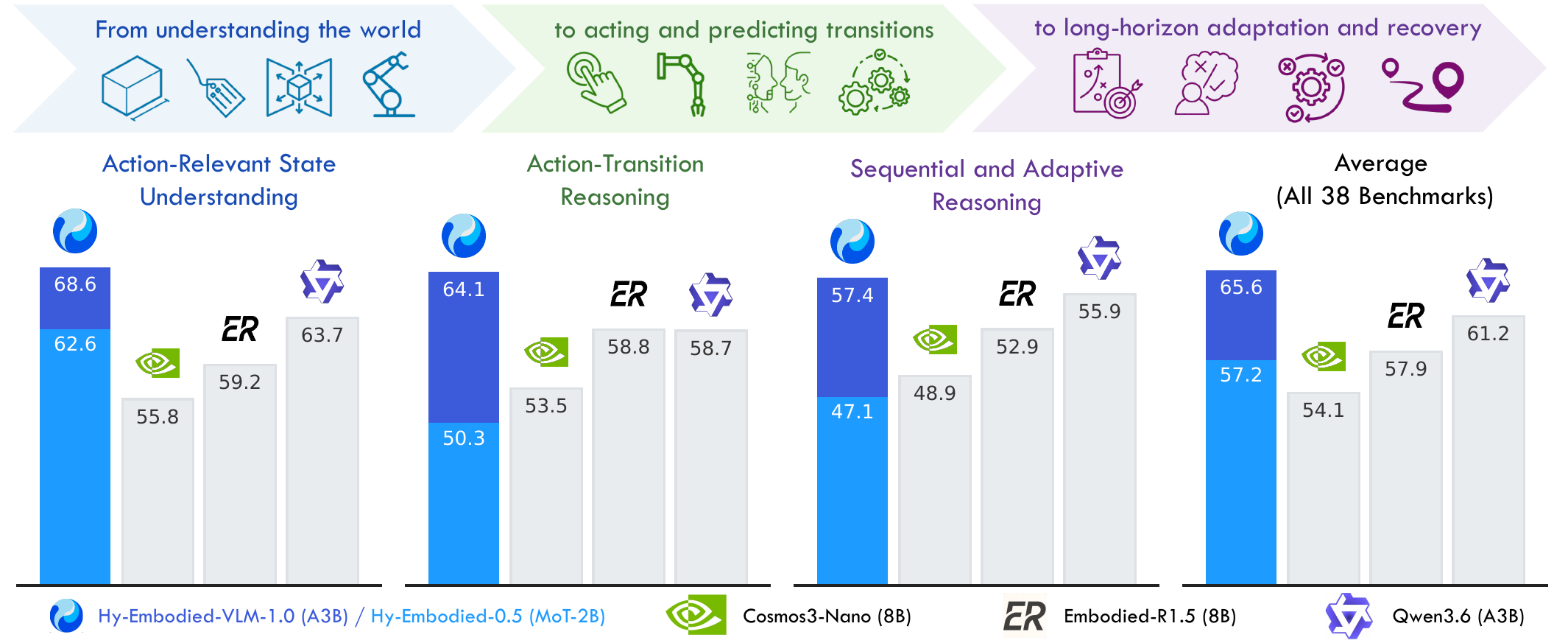}
  \vspace{-5pt}
  \caption{\textbf{Performance Comparison.}
Hy-Embodied-VLM-1.0 consistently outperforms strong VLM baselines of comparable scale across the three capability levels of our embodied intelligence taxonomy, achieving the best overall performance across all 38 benchmarks.}
  \label{fig:1-teaser}
\end{figure}

\newpage
\tableofcontents
\newpage

\section{Introduction}

Agents operating in the digital world have made remarkable progress in recent years~\citep{gemini3,openai2025gpt5,anthropic2025claude4,deepseek2024v3,park2023generative,yang2024sweagent,xie2024osworld}, exhibiting increasingly strong capabilities for multi-step interaction, tool use, and long-horizon reasoning. These advances have enabled agents to autonomously pursue complex goals through iterative perception, reasoning, and action. Extending such agentic capabilities to the physical world, however, requires more than fundamental multimodal perception, understanding, and reasoning~\citep{liu2024llava,bai2023qwenvl}. Models need to reason about and execute actions, understand how the environment evolves in response to their actions, and continuously adapt their behavior to changing physical conditions. Developing these capabilities remains a significant challenge.
To address these challenges and build generalizable, high-performing agents for the physical world, we develop an action-centric capability taxonomy consisting of three progressive levels. \textbf{Action-Relevant State Understanding} focuses on accurately and robustly understanding the states of the agent and its environment, including fundamental physical properties, three-dimensional spatial structures, and embodiment-related attributes. \textbf{Action–Transition Reasoning} focuses on understanding actions and agent–environment interactions, planning basic actions, and reasoning about their consequences and the underlying physical principles. \textbf{Sequential and Adaptive Reasoning} further extends these capabilities to complex, long-horizon action sequences, requiring models to reason over past actions and future goals and to perform reflection, repair, and recovery during complex interactions.

In this report, we introduce Hy-Embodied-VLM-1.0, an efficient and powerful embodied foundation model specifically designed for embodied agents operating in the physical world. Hy-Embodied-VLM-1.0 continues the central vision of the Hy-Embodied series: transferring the broad intelligence developed in the digital world to agents that perceive, reason, and act in the physical world. Following the Vision-Language Model paradigm~\citep{liu2024llava}, we leverage the extensive knowledge and reasoning capabilities of language models, retain strong multimodal perception and understanding, and extend these capabilities toward action-centric reasoning and interaction in physical environments. As a substantial upgrade over Hy-Embodied-0.5, Hy-Embodied-VLM-1.0 introduces major improvements in both data and model design. On the data side, our action-centric capability taxonomy guides the construction of a systematic data pipeline and carefully designed mixtures across pre-training and post-training. On the model side, we integrate the Hy3-A3B language backbone with the Hy-ViT2 vision encoder~\citep{team2025hunyuanocr,yu2026viq} and adopt an efficient Mixture-of-Experts~\citep{moe} architecture, enabling strong physical-world understanding and interaction capabilities while maintaining high inference efficiency for latency-sensitive deployment.

Evaluation continues to play a central role in the development of Hy-Embodied-VLM-1.0. To more comprehensively assess the capabilities required by physical-world agents, we substantially expand our evaluation suite to 38 benchmarks. These benchmarks span three progressive capability levels, ranging from fine-grained multimodal and environmental perception, through action and interaction understanding, to complex agentic tasks requiring sequential decision-making and long-horizon reasoning. We compare Hy-Embodied-VLM-1.0 with three recent state-of-the-art models of comparable scale: Qwen3.6-A3B~\citep{qwen36_35b_a3b}, NVIDIA Cosmos 3-8B~\citep{agarwal2026cosmos3}, and Embodied-R1.5-8B~\citep{yuan2026embodied-r15}. Among the compared models, Hy-Embodied-VLM-1.0 ranks first on 19 of the 38 benchmarks and second on another 11. It also achieves the highest average performance across all three capability categories, outperforming the strongest competing model, Qwen3.6-A3B, by 4.4\% on average. Compared with the previous-generation Hy-Embodied-0.5 MoT-2B, our model improves average performance by 8.4\%. Moreover, with only 3B activated parameters—approximately one-tenth of those used by the previous-generation A32B model—it achieves nearly comparable overall performance. These results demonstrate that Hy-Embodied-VLM-1.0 delivers substantial gains in both capability and computational efficiency. We further assess the model on vision-and-language navigation tasks that place greater demands on systematic interaction and long-horizon decision-making. On the R2R-CE benchmark~\citep{anderson2018vision,krantz_vlnce_2020}, our model achieves state-of-the-art performance among methods in the RGB-only setting. It also exhibits strong zero-shot navigation capabilities on the Matterport3D Object Goal Navigation task~\citep{chang2017matterport3d}, demonstrating its ability to transfer physical-world understanding and agentic reasoning to challenging interactive environments.
\section{Data and Taxonomy} \label{data}

\subsection{Taxonomy}

Embodied foundation models are expected to operate in physical environments where perception, reasoning, and decision-making are tightly coupled. Unlike general-purpose VLMs, which primarily answer visual questions or describe image content, embodied models must understand the physical world in an action-relevant manner. This requires recognizing entities and their properties, reasoning about spatial and robotic constraints, understanding interactions and potential state changes, and supporting temporally extended decision-making under goals and feedback. To systematically characterize these requirements, we introduce a capability taxonomy for Hy-Embodied-VLM-1.0.

\begin{figure}[tb]
  \centering
  \includegraphics[width=\linewidth]{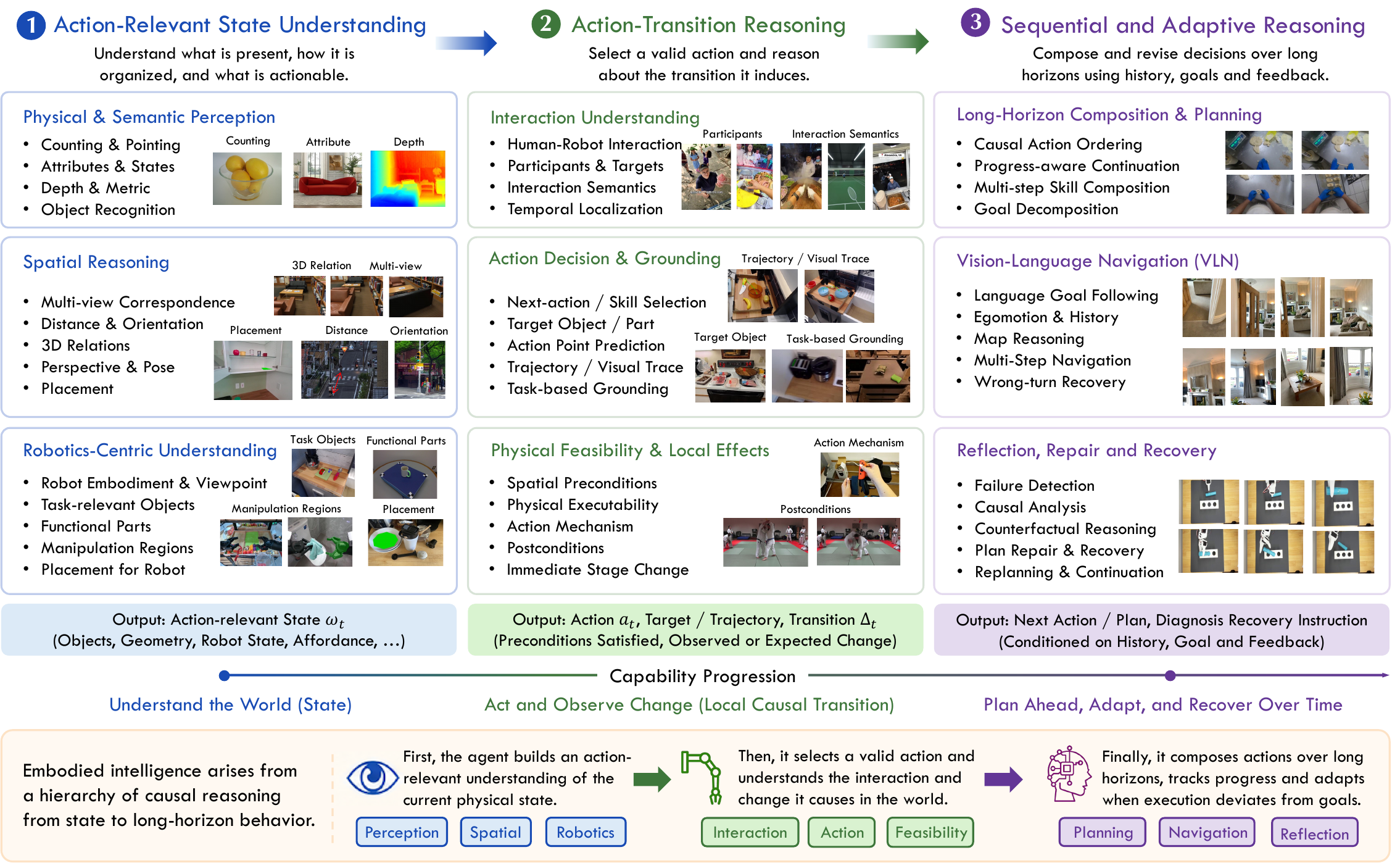}
  \caption{\textbf{Capability Taxonomy of Hy-Embodied-VLM-1.0.} We organize embodied intelligence into a three-level hierarchy that progresses from action-relevant state understanding, to action-transition reasoning, and finally to sequential and adaptive reasoning. The first level builds an actionable representation of the current physical scene; the second level grounds immediate actions and reasons about the local transitions they induce; and the third level composes and revises decisions over time through planning, navigation, reflection, and recovery. This taxonomy guides our data construction and benchmark evaluation, linking perception, action, and feedback into a unified capability progression.}
  \label{fig:2-taxonomy}
\end{figure}

As illustrated in Figure~\ref{fig:2-taxonomy}, the taxonomy organizes embodied capabilities into three conceptual levels: \textit{Action-Relevant State Understanding}, \textit{Action-Transition Reasoning}, and \textit{Sequential and Adaptive Reasoning}. These levels describe a broad progression of capabilities, from understanding the current physical state, through reasoning about actions and local state transitions, to planning, navigation, and recovery over extended horizons. Rather than prescribing a fixed inference procedure, the taxonomy serves as a unified framework that connects capability formulation with targeted data construction and benchmark evaluation, ensuring consistency across model development and empirical assessment.

\paragrapha{Action-Relevant State Understanding.}
The first level focuses on understanding the current physical scene in an action-relevant manner. 
For embodied agents, perception is not limited to object recognition or image captioning; it should provide useful information for downstream interaction, such as objects, parts, attributes, states, geometry, robot-centric cues, affordances, and task-relevant conditions. 
We decompose this level into three complementary capabilities: physical and semantic perception, spatial reasoning, and robotics-centric understanding.

\textit{Physical and Semantic Perception} provides the basic visual grounding needed for embodied reasoning. 
It includes object and part recognition, counting and pointing, attribute and state understanding, as well as depth and metric estimation. 
These capabilities allow the model to identify what is present in the scene and describe the physical or semantic properties that may affect later decisions.

\textit{Spatial Reasoning} extends visual perception to the geometric structure of the environment. 
It covers 3D geometry and relations, metric distance and orientation, perspective taking, pose reasoning, multi-view correspondence, spatial memory, free-space and placement reasoning, and scene reconstruction. 
These abilities help the model understand where entities are located, how they are organized, and how spatial structure changes or remains consistent across viewpoints and observations.

\textit{Robotics-Centric Understanding} further connects scene understanding to robot operation. 
It emphasizes the robot's embodiment and viewpoint, task-relevant objects, functional parts, affordances, manipulation regions, and feasible placement regions. 
This capability is important because an embodied model must not only understand the scene as an observer, but also identify which objects, regions, and constraints are relevant for a robot to act.

Together, these capabilities characterize the model's ability to form action-relevant scene understanding. 
They provide the foundation for downstream embodied reasoning by grounding objects, geometry, robotic context, affordances, and physical constraints in the current environment.

\paragrapha{Action-Transition Reasoning.}
The second level focuses on reasoning about actions, interactions, and the local changes associated with them. 
While state understanding describes what is currently true in the environment, embodied decision-making also requires the model to understand what can be done, where an action should be applied, and how the scene may change after an action or interaction. 
We organize this level into interaction understanding, action decision and grounding, and physical feasibility with local effects.

\textit{Interaction Understanding} captures the ability to recognize and interpret interactions among humans, robots, and objects. 
It includes identifying participants and targets, understanding interaction semantics, and localizing interactions over time. 
This capability is especially important for embodied scenarios where the model must reason about ongoing human-object, robot-object, or human-robot activities rather than isolated static objects.

\textit{Action Decision and Grounding} concerns immediate action-oriented reasoning. 
Given a scene and a task context, the model should be able to identify plausible next actions or skills, ground them to target objects or functional parts, predict action points, generate trajectories or visual traces, and support task-conditioned grounding for manipulation or navigation. 
This capability connects scene understanding with concrete embodied decisions.

\textit{Physical Feasibility and Local Effects} emphasizes whether an action is physically meaningful under the current conditions and what local consequences it may induce. 
It includes reasoning about spatial preconditions, physical executability, action mechanisms, postconditions, and immediate state changes. 
For example, an embodied model should be able to reason about whether an object can be grasped, where the action should be applied, and what visible or functional change may follow.

Overall, Action-Transition Reasoning captures the model's ability to connect the current scene with possible actions and their local consequences. 
It bridges static understanding and temporally extended behavior by emphasizing interaction, grounding, feasibility, and short-range physical change.

\paragrapha{Sequential and Adaptive Reasoning.}
The third level focuses on embodied reasoning over time. 
Many real-world tasks cannot be solved through a single observation or a single local action: they require maintaining history, following language goals, composing multiple steps, monitoring progress, and adapting when execution deviates from expectation. 
We therefore define Sequential and Adaptive Reasoning to cover long-horizon composition and planning, vision-language navigation, and reflection-based repair and recovery.

\textit{Long-Horizon Composition and Planning} requires the model to organize multiple actions or subgoals into a coherent temporal structure. 
It includes goal decomposition, causal action ordering, progress-aware continuation, and multi-step skill composition. 
Rather than only selecting a locally plausible next action, this capability emphasizes whether the model can reason about how a sequence of decisions contributes to a longer-term objective.

\textit{Vision-Language Navigation} represents a class of temporally extended embodied tasks where the model must follow language goals through continuous visual observations. 
It involves language goal following, spatial memory, egomotion and history, viewpoint and map reasoning, multi-step navigation, and wrong-turn recovery. 
This capability highlights the need to combine instruction grounding, spatial reasoning, and sequential decision-making under partial and changing observations.

\textit{Reflection, Repair, and Recovery} focuses on adaptation when execution or planning does not proceed as intended. 
It includes failure detection and diagnosis, causal analysis, counterfactual reasoning, plan repair, recovery, replanning, and continuation. 
Such capabilities are essential for robust embodied behavior, since real-world environments are uncertain and physical execution often introduces unexpected outcomes.

Together, these capabilities characterize the model's ability to reason over history, goals, and feedback in temporally extended tasks. 
They move beyond local action reasoning toward planning, navigation, progress tracking, and recovery, enabling a broader form of embodied intelligence that can adapt over time.

In summary, the proposed taxonomy frames embodied intelligence as a hierarchy of capabilities from action-relevant state understanding, to action-transition reasoning, and finally to sequential and adaptive reasoning. 
This capability hierarchy provides a top-down design principle for Hy-Embodied-VLM-1.0: it guides how we organize training data, how we select and group evaluation benchmarks, and how we analyze the strengths and limitations of the resulting model across different forms of physical-world reasoning.

\subsection{Data}

\subsubsection{Pre-training Data}

Hy-Embodied-VLM-1.0 inherits the pre-training data mixture from Hy-Embodied-0.5~\citep{team2026hy-emb-05} without additional modification. 
This pre-training corpus provides a broad foundation for physical-world understanding, covering visual perception, spatial reasoning, embodied-centric supervision, and general vision-language understanding. 
Specifically, it includes large-scale data for object grounding, detection, depth estimation, segmentation, pointing and counting, spatial correspondence, geometry, affordance, trajectory understanding, and planning-oriented reasoning. 
We keep this stage unchanged because it already establishes the low-level visual grounding and general embodied understanding required by our capability taxonomy. 
Therefore, the main data evolution of Hy-Embodied-VLM-1.0 is introduced in the subsequent supervised fine-tuning and reinforcement learning stages, where we further specialize the model toward action-transition reasoning and sequential adaptive reasoning.

\subsubsection{Supervised Fine-tuning Data}

\begin{figure}[tb]
  \centering
  \includegraphics[width=\linewidth]{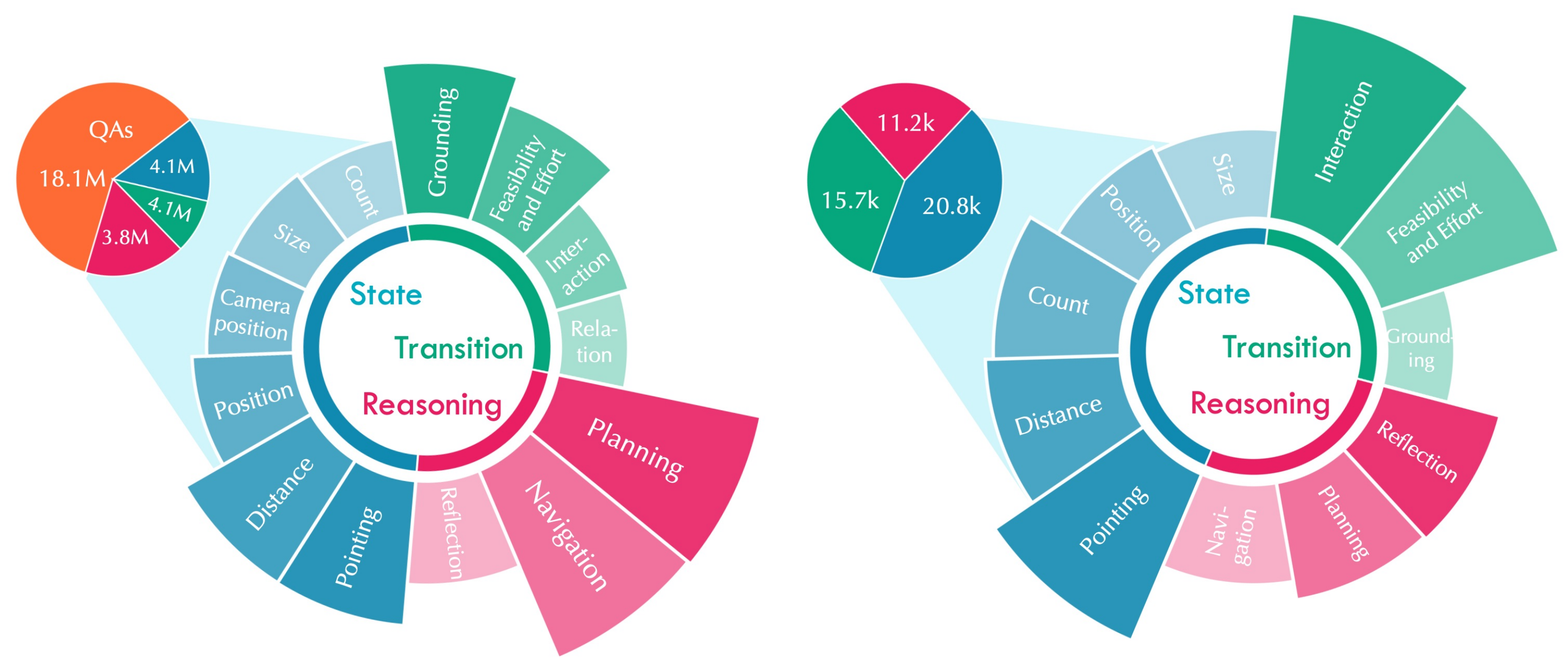}
  \caption{\textbf{Data Distribution for Supervised Fine-tuning and Reinforcement Learning.} We organize the post-training data of Hy-Embodied-VLM-1.0 according to the proposed capability taxonomy, covering state understanding, transition reasoning, and sequential reasoning. The SFT data expands the inherited Hy-Embodied-0.5 mixture with additional capability-oriented supervision for depth reasoning, task-conditioned grounding, social interaction, trajectory reasoning, physical-causal reasoning, failure-aware reasoning, and vision-language navigation. The RL mixture focuses on verifiable embodied reasoning tasks, providing reward-driven supervision across state reasoning, grounding, interaction, planning, reflection, and navigation-related capabilities.}
  \label{fig:3-data}
\end{figure}

The supervised fine-tuning stage further specializes Hy-Embodied-VLM-1.0 toward the capability hierarchy introduced above. 
We inherit the high-quality SFT data mixture from Hy-Embodied-0.5, which already covers embodied perception, spatial understanding, grounding, affordance, trajectory, and planning-oriented reasoning~\citep{team2026hy-emb-05}. 
On top of this foundation, we introduce additional capability-oriented data to better cover the three levels of our taxonomy: action-relevant state understanding, action-transition reasoning, and sequential and adaptive reasoning. 
As illustrated in Figure~\ref{fig:3-data}, the SFT mixture expands state-level supervision with depth reasoning and task-conditioned grounding, transition-level supervision with social interaction, trajectory reasoning, and physical-causal reasoning, and sequential-level supervision with failure-aware reasoning and vision-language navigation. 
We also adapt selected embodied manipulation data from Embodied-R1.5~\citep{yuan2026embodied-r15}, while reorganizing them according to our capability taxonomy rather than treating them as a separate data source. 
All added data are constructed to strengthen general embodied capabilities, rather than to target any individual benchmark.

\paragrapha{Depth Reasoning.}
To improve metric depth understanding and geometry-aware reasoning across indoor and outdoor scenes, we construct supervision from Argoverse~2~\citep{wilson2023argoverse}, Matterport3D~\citep{chang2017matterport3d}, nuScenes~\citep{caesar2020nuscenes}, ScanNet++~\citep{yeshwanth2023scannet++}, and Taskonomy~\citep{zamir2018taskonomy}. For aligned RGB--D data, we obtain metric depth directly from valid depth pixels. For driving data, we transform calibrated LiDAR points into the camera frame, project them onto the image plane, filter invalid projections, and randomly retain up to 100 candidates per image. We normalize the reference focal length to 1,000 pixels through isotropic resizing, updating the intrinsics and point coordinates accordingly, and rectify images when required by their calibration. Each target is sampled at least 60 pixels from the image boundary and indicated by a red arrow ending at a solid green anchor, providing localized metric-depth supervision. Based on these geometrically grounded annotations, we further synthesize chain-of-thought (CoT) examples that require combining multiple depth cues through intermediate reasoning steps. For evaluation, we curate Depth-InHouse benchmark, an in-house 4,677-question multiple-choice benchmark for depth-dependent real-world spatial understanding, including counterintuitive cases designed to distinguish genuine depth reasoning from shortcuts based on 2D appearance.

\paragrapha{Task-conditioned Grounding and Affordance Localization.}
To strengthen action-relevant state understanding and action grounding in manipulation scenarios, we incorporate data that requires the model to localize task-relevant objects, functional parts, feasible interaction regions, and target placement regions under natural-language instructions.
Unlike generic referring expression grounding, these samples emphasize whether the grounded region is meaningful for a potential robot action. 
For example, the model may need to identify a graspable part, a pullable surface, a suction point, or a placement region that satisfies the task context and physical constraints. 
This type of data supports both robotics-centric state understanding and action-transition reasoning, since the model must connect visual entities with their functional roles in embodied tasks.

\paragrapha{Social Interaction.}
To enhance interaction understanding, we introduce data that requires the model to recognize participants, targets, interaction semantics, and temporally localized social or human-object activities. The corpus integrates three complementary data sources. The primary source comprises egocentric videos of human-human social interactions, collected from public web videos and our in-house recordings. For these videos, we manually annotate the social cues of the person currently interacting with the ego wearer, covering interactant identification, gaze and head-direction estimation, pointing gesture and target recognition, body pose and orientation understanding, and emotion recognition. These annotations are provided in two complementary supervision formats: structured JSON predictions and targeted question-answering pairs, which together form the core of the dataset. To broaden the coverage of social interactions, we further incorporate publicly available data sampled from AVA v2.2~\citep{gu2018ava} and HICO-DET~\citep{chao2018learning}, constructing question-answering pairs that cover human-centric motion, social behaviors, object manipulation, and human-object interactions. These samples require the model to reason over temporal context, localize relevant humans and objects when necessary, and identify the corresponding interaction patterns. In our implementation, all coordinates are normalized to integers ranging from 0 to 1000 and represented in a fixed output format.

\paragrapha{Object-centric and Robot-centric Trajectory Supervision.}
To enhance action-transition reasoning, we introduce trajectory-oriented supervision that teaches the model how objects, robot end-effectors, or human-operated objects move during manipulation. 
These data present a visual observation together with a task or motion description, and supervise the model with ordered point sequences representing object motion, end-effector motion, or human-demonstrated manipulation traces. 
Compared with static point grounding, trajectory supervision exposes the model to the spatial structure of action execution, helping it reason about direction, continuity, intermediate waypoints, and local state changes caused by manipulation. 
These data mainly strengthen action decision and grounding, physical transition understanding, and the connection between local actions and their expected visual evolution.

\paragrapha{Causal Reasoning.}
To strengthen physically grounded causal reasoning, we construct supervision from egocentric and robotic manipulation videos~\citep{grauman2022ego4d,lin2022egocentric,sermanet2024robovqa,bu2025agibot_iros}, building on prior work on causal task understanding and procedure planning~\citep{jia2022egotaskqa,qiu2024egoplan,azzolini2025cosmos,lu2026token,zhao2026gem}.
For each video, we first identify the overall task goal and decompose the procedure into ordered steps with explicit preconditions, effects, and inter-step dependencies.
Each step is then temporally grounded to a visually supported interval, while ambiguous or unsupported steps are removed.
Within each retained interval, we track the agent, objects, and state change, using keyframes to anchor the transition. Complex actions are decomposed into timestamped atomic interactions.
These structured annotations are paired with task-appropriate visual evidence and converted into question-answering samples covering action executability, effect prediction, step composition, and causal robustness.
Reasoning traces are generated under physical constraints, and all samples are filtered for visual grounding and logical consistency.

\paragrapha{Failure-aware Robotic Reasoning.}
To strengthen reflection, repair, and recovery, we include failure-aware robotic reasoning data from robot execution videos, perturbed task plans, and failure cases.
These samples ask the model to judge execution status, verify whether a subtask plan is reasonable, identify erroneous actions or stages, and provide recovery-oriented suggestions when appropriate. 
Such supervision encourages the model to reason about execution progress, causal failure patterns, and corrective behavior from visual and textual context. 
This data is especially relevant to sequential and adaptive reasoning, where the model must go beyond recognizing the current scene and instead reason about whether the task is proceeding as intended and how it may be repaired after deviation.

\paragrapha{Vision-Language Navigation.}
We construct a two-stage vision-language navigation (VLN) dataset to enhance instruction following, spatial memory, and closed-loop decision-making in indoor environments.
Trajectories are collected in Habitat~\citep{savva2019habitat} on the R2R-CE \texttt{train} split~\citep{anderson2018vision,krantz_vlnce_2020} using only egocentric RGB observations.
In the first stage, a shortest-path oracle policy generates expert demonstrations by executing low-level actions along geodesic shortest paths.
At each visited state, the language instruction and a temporally sampled observation history are paired with a target action chunk containing four future oracle actions from the discrete action space \{\texttt{forward}, \texttt{left}, \texttt{right}, \texttt{stop}\}.
As expert demonstrations provide limited coverage of the state distribution induced by the learned policy, the second stage applies Dataset Aggregation (DAgger)~\citep{ross2011reduction} using the first-stage policy.
Following Efficient-VLN~\citep{zheng2025efficientvln}, the behavior policy selects the oracle policy with probability $\beta_t=1-\alpha^{t/T}$ and the learned policy otherwise, where $t$ denotes the current step, $T$ denotes the ground-truth path length in steps, and $\alpha=0.5$.
This schedule progressively shifts execution from learned-policy exploration to oracle guidance, exposing the model to error-induced states while constraining rollout length.
For every visited state, the corresponding oracle action chunk is retained as the supervision target irrespective of the policy used for execution, providing correction signals for direction-selection errors, premature or missed termination, and accumulated trajectory deviations.
After collection, we remove malformed and post-termination samples, truncate repetitive action sequences, and rebalance turning and stopping examples to improve goal-state recognition and termination prediction.
%

\subsubsection{Reinforcement Learning Data}

The reinforcement learning stage further improves Hy-Embodied-VLM-1.0 on verifiable embodied reasoning tasks. 
Compared with supervised fine-tuning, which provides broad capability coverage through high-quality demonstrations, the RL stage emphasizes tasks whose outputs can be automatically parsed and evaluated with reliable rule-based rewards. 
As illustrated in Figure~\ref{fig:3-data}, our RL data follows the same capability-oriented organization as the SFT mixture, covering state reasoning, transition reasoning, pointing and grounding, planning, reflection, and navigation-related capabilities. 
The training mixture contains both our existing embodied RL data and selected additional data adapted from Embodied-R1.5~\citep{yuan2026embodied-r15}, which are reorganized according to our taxonomy to strengthen spatial reasoning, object-centric motion understanding, point localization, and 3D trajectory reasoning. 
All RL samples are constructed around general embodied capabilities rather than individual benchmarks.

\paragrapha{Spatial and Metric State Reasoning.}
To strengthen action-relevant state understanding, we include multi-frame spatial reasoning data that requires the model to estimate physical quantities and reason about spatial relations in indoor environments. 
These tasks cover object size estimation, object counting, absolute distance estimation between objects, room size estimation, and navigation-oriented spatial relation reasoning. 
The inputs are represented as multiple visual observations from the same environment, encouraging the model to integrate information across views and build a more stable understanding of scale, distance, layout, and orientation. 
Such data provides reward-driven supervision for metric physical reasoning and spatially grounded scene understanding.

\paragrapha{Pointing, Grounding, and Affordance Prediction.}
To improve robotics-centric grounding and action-oriented localization, we include tasks that require the model to output structured coordinates or regions, such as points, bounding boxes, and missing-object locations. 
These samples train the model to identify robot-relevant regions, affordance areas, contact points, placement positions, and geometrically implied target locations under task or scene constraints. 
Compared with open-ended language supervision, these tasks provide more direct learning signals for embodied grounding because the predicted spatial outputs can be evaluated against ground-truth coordinates or regions. 
They mainly strengthen the connection between visual-language understanding and actionable spatial representations.

\paragrapha{Object-centric and Robot-centric Trajectory Reasoning.}
To enhance action-transition reasoning, we include trajectory prediction data that requires the model to output ordered point sequences describing object motion, robot end-effector motion, or articulated-part movement. 
The 2D trajectory data encourages the model to reason about how manipulated objects or robot embodiments move in the image plane, while the 3D trace data further requires depth-aware trajectory prediction for articulated objects such as doors and handles. 
These samples expose the model to the spatial continuity of action execution and the local changes induced by manipulation. 
They provide structured supervision for reasoning about action direction, motion topology, intermediate waypoints, and visual consequences of embodied actions.

\paragrapha{Sequential Process and Navigation Reasoning.}
To improve sequential and adaptive reasoning, we include tasks involving process ordering, navigation direction prediction, and history-conditioned spatial decisions. 
These samples require the model to reason beyond a single static observation by comparing multiple frames, identifying plausible temporal order, or deciding how an embodied agent should move relative to a target. 
Although these tasks do not define a complete closed-loop system, they provide verifiable signals for temporal organization, navigation-oriented reasoning, and progress-aware embodied understanding.

\section{Model Architecture}
\label{arch}

We build our vision-language model upon the Hy-series by adopting the A3B large language model as the language backbone. The backbone follows a Mixture-of-Experts (MoE) architecture~\citep{moe} with approximately 30B total parameters, while only around 3B parameters are activated for each token during inference. Following the standard vision-language architecture~\citep{liu2024llava}, a dedicated vision encoder is coupled with the language model through a lightweight connector. This modular design allows the vision encoder to specialize in visual perception while the language model focuses on reasoning and text generation. Instead of introducing complex architectural modifications, we adopt well-established components to ensure stable training and practical deployment.

A key objective of our design is to achieve a favorable balance between model capability and inference efficiency. Previous studies and empirical experience consistently demonstrate that scaling language models leads to stronger reasoning ability, richer expression, and better instruction following. However, many existing large language models employ dense architectures in which all model parameters participate in every forward pass, resulting in substantial computational cost during inference. To address this issue, A3B adopts a MoE architecture that dynamically routes each token to a small subset of specialized experts. Consequently, only 3B parameters are activated during inference, enabling the model to retain the representational capacity of a much larger model while maintaining computational cost comparable to that of a significantly smaller one. This property makes the model particularly suitable for large-scale deployment.

On the vision side, we employ a native-resolution Vision Transformer (ViT)~\citep{Dosovitskiyetal2020vit,Dehghanietal2023navit,tschannen2025siglip} as the visual encoder, following the design philosophy of the Hy-ViT series~\citep{team2025hunyuanocr,yu2026viq}. Unlike conventional vision encoders that require every image to be resized to a fixed resolution, the deployed Hy-ViT2 encoder accepts images with arbitrary resolutions and aspect ratios. This capability preserves fine-grained visual information and reduces the distortion caused by image resizing. The extracted visual features are projected into the representation space of the language model through the connector and are processed jointly with the textual context within a unified sequence. We further introduce several multimodal adaptations to improve the modeling of visual inputs while preserving the overall simplicity and effectiveness of the architecture.

\section{Training}
\label{sec:training}

\begin{figure}[t]
\centering
\includegraphics[width=0.9\linewidth]{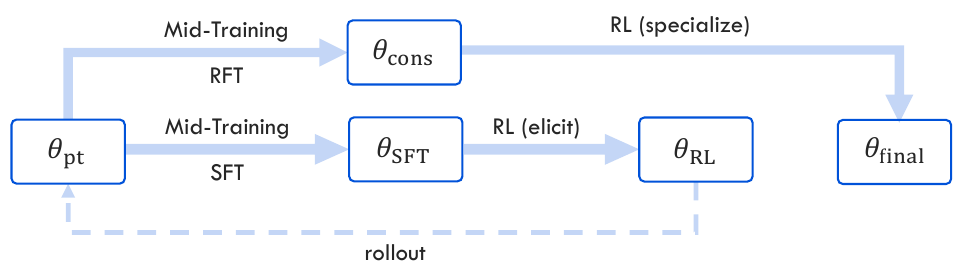}
\caption{Training pipeline of Hy-Embodied-VLM-1.0. Supervised fine-tuning and RL turn $\theta_{\mathrm{pt}}$ into a reasoner $\theta_{\mathrm{rl}}$, whose rollouts are used to re-train a fresh model $\theta_{\mathrm{cons}}$ from $\theta_{\mathrm{pt}}$ by rejection-sampling fine-tuning; a final reward-specialized RL stage yields the deployed model $\theta_{\mathrm{final}}$.}
\label{fig:training-pipeline}
\end{figure}

The training of Hy-Embodied-VLM-1.0 is organized around a single goal: to cultivate \emph{embodied agentic reasoning}, so that the model does not merely produce embodied answers but reasons about states, actions, transitions, and long-horizon decisions in the sense of the capability taxonomy. Building on a pre-trained VLM, we inject physical-world competence through embodied pre-training and supervised fine-tuning, and then develop reasoning through a self-evolving post-training loop that couples reinforcement learning with rejection-sampling fine-tuning. All stages share the Hy3-A3B Mixture-of-Experts backbone and the Hy-ViT2 vision encoder, activating roughly $3$B parameters per token. Figure~\ref{fig:training-pipeline} illustrates the overall training path: reinforcement learning first elicits a strong reasoner $\theta_{\mathrm{rl}}$ from the supervised checkpoint $\theta_{\mathrm{sft}}$, whose rollouts are rejection-sampled and folded back to re-train a fresh model $\theta_{\mathrm{cons}}$ from the pre-trained checkpoint $\theta_{\mathrm{pt}}$, which a final reward-specialized RL stage then turns into the deployed agent $\theta_{\mathrm{final}}$.

\subsection{Pre-Training}
\label{sec:pretraining}

Hy-Embodied-VLM-1.0 is initialized from a pre-trained VLM with robust general vision-language alignment, which lets us reuse broad multimodal knowledge and devote the pre-training budget to the physical-world competencies that general VLMs typically lack. On this initialization we perform embodied continued pre-training over a large multimodal corpus that couples low-level visual perception (2D/3D grounding, detection, depth estimation, segmentation, pointing, and counting) with spatial correspondence, geometry, affordance, and trajectory and planning-oriented supervision. Optimized with the standard autoregressive objective, this stage establishes the fine-grained visual grounding and general embodied understanding required by the lowest level of our taxonomy, and yields the foundational checkpoint $\theta_{\mathrm{pt}}$ that anchors all subsequent post-training.

\subsection{Mid-Training \& SFT}
\label{sec:sft}

Between pre-training and the post-training loop, we run a mid-training and supervised fine-tuning stage that turns the perceptually grounded backbone into a broadly capable instruction follower covering the full capability taxonomy, while lightly introducing an explicit reasoning mode. We treat this as a single continuous process over a curated mixture rather than two sharply delimited phases, gradually shifting from broad capability coverage toward more targeted, higher-quality supervision as training proceeds.

Concretely, the mixture combines a large amount of general embodied supervision with a small, carefully curated set of high-quality \emph{thinking} traces. The general supervision is spread across all three taxonomy levels of Section~\ref{data} so that no capability is starved, ranging from physical and semantic perception, depth and metric estimation, and spatial and robotics-centric grounding at the state level, through interaction understanding, action grounding, and physical feasibility at the transition level, to long-horizon planning, failure-aware reflection, and vision-language navigation at the sequential level. This breadth gives the model near-complete coverage of the embodied skills we target and teaches it the output interfaces required for downstream reward-driven training. On top of it we mix in the thinking traces, i.e., long chain-of-thought demonstrations that externalize intermediate reasoning before committing to an answer. These traces are produced by a human--model collaborative pipeline and individually verified for reasoning quality, logical correctness, and answer consistency. This thinking set is kept intentionally small, on the order of a few tens of thousands of examples: its purpose is not to teach reasoning content but to \emph{activate} a reliable thinking format and provide a cold start for RL, while the depth and breadth of embodied reasoning are cultivated by the self-evolving RL/RFT loop that follows. The resulting checkpoint
$\theta_{\mathrm{sft}}$ is thus a broadly competent policy carrying a lightly seeded reasoning behavior.

\subsection{Reinforcement Learning}
\label{sec:rl}

Reinforcement learning is the engine of embodied agentic reasoning in our pipeline, and we apply it at two points: an \emph{elicitation} stage that turns the shallowly seeded SFT policy into a genuine reasoner (Section~\ref{sec:rl-elicit}), and a \emph{specialization} stage that, after reasoning has been consolidated by RFT, sharpens precision by training
reward-specialized policies that are then fused into the deployed model
(Section~\ref{sec:rl-merge}). Both stages share the same
optimizer, reward family, and normalization scheme, which we describe first. Embodied targets
are highly heterogeneous. They span geometric grounding, trajectory prediction, discrete
decisions, continuous metric estimation, and open-ended reasoning, so the central technical
challenge is to provide a low-noise and comparable learning signal across all of them.

\subsubsection{Policy Optimization with GRPO}
\label{sec:grpo}

We optimize the policy with GRPO\citep{shao2024grpo}, which avoids a separately learned value network by estimating advantages from a group of sampled responses. For each multimodal input $x=(I,Q)$, we draw a group of $G$ responses $\{y_1,\dots,y_G\}$ from the behavior policy $\pi_{\theta_{\mathrm{old}}}$ and score each with the task-aware reward described below, yielding $\mathbf{r}=\{r_1,\dots,r_G\}$. The policy is updated with a token-level clipped surrogate
objective:
\begin{equation}
\label{eq:grpo}
\mathcal{L}_{\mathrm{RL}}(x)
= -\frac{1}{\sum_{i=1}^{G}|y_i|}
\sum_{i=1}^{G}\sum_{t=1}^{|y_i|}
\min\!\Big(
\rho_{i,t}\,\hat{A}_i,\;
\mathrm{clip}\big(\rho_{i,t},\,1-\epsilon_{\mathrm{low}},\,1+\epsilon_{\mathrm{high}}\big)\hat{A}_i
\Big),
\end{equation}
where the per-token importance ratio is
\begin{equation}
\label{eq:ratio}
\rho_{i,t}
= \frac{\pi_{\theta}(y_{i,t}\mid x, y_{i,<t})}
       {\pi_{\theta_{\mathrm{old}}}(y_{i,t}\mid x, y_{i,<t})} .
\end{equation}
The advantage $\hat{A}_i$ is shared across all tokens of rollout $y_i$. In its standard form, GRPO computes a group-relative advantage by normalizing rewards within the sampled group,
\begin{equation}
\label{eq:group-adv}
\hat{A}_i^{\mathrm{group}}
= \frac{r_i - \mu(\mathbf{r})}{\sigma(\mathbf{r}) + \epsilon},
\qquad
\mu(\mathbf{r})=\tfrac{1}{G}\textstyle\sum_{j} r_j,\;\;
\sigma(\mathbf{r})=\sqrt{\tfrac{1}{G}\textstyle\sum_{j}\big(r_j-\mu(\mathbf{r})\big)^2}.
\end{equation}
We use a group size of $G=16$ and an essentially on-policy update scheme in which the PPO mini-batch size matches the rollout batch size \citep{schulman2017ppo}. We further adopt asymmetric clipping with an effective importance-ratio range of $[0.8,1.35]$, which we find more stable than symmetric clipping for long-chain multimodal RL, and we mask groups with zero
reward variance ($\sigma(\mathbf{r})=0$) since they carry no relative learning signal.

Because our task mixture is highly heterogeneous, the per-group standard deviation in Eq.~\eqref{eq:group-adv} leaves reward scales incomparable across tasks. Building on prior work on stabilizing GRPO \citep{yuan2026embodied-r15,li2026tempsamp}, we therefore keep the within-group mean but normalize by the reward standard deviation over the whole batch $\mathcal{B}$,
\begin{equation}
\label{eq:global-adv}
\hat{A}_i = \frac{r_i - \mu_{\mathrm{group}}}{\sigma_{\mathcal{B}} + \epsilon},
\end{equation}
which unifies gradient magnitudes across capabilities. Both RL stages below use $\hat{A}_i$ of Eq.~\eqref{eq:global-adv} in the objective of Eq.~\eqref{eq:grpo}.

\subsubsection{Task-Aware Embodied Rewards}
\label{sec:reward}

A single uniform reward is inadequate for embodied outputs. We therefore assign each sample a reward function according to the structure of its target, so that partial spatial or temporal correctness is distinguished from complete failure whenever this is meaningful:
\begin{equation}
\label{eq:reward-general}
r = R_{\tau}(y, y^{\star}) \in [0,1],
\end{equation}
where $\tau$ is the task type, $y$ the model response, and $y^{\star}$ the reference. Our principle is to use deterministic, structure-aware rewards whenever the target admits reliable parsing, and to fall back to an LLM judge only when necessary.

\paragraph{Geometric grounding.} For tasks with explicit geometric structure we use dense overlap- or distance-based rewards rather than exact match. A single bounding box is scored by Intersection-over-Union, $R = \mathrm{IoU}(b,b^{\star})$. For multi-object detection we first solve a Hungarian assignment between predicted and reference boxes on their center distances, yielding a matched set $\mathcal{M}$, and then penalize count mismatch,
\begin{equation}
\label{eq:multibox}
R = \Bigg(\frac{1}{|\mathcal{M}|}\sum_{(i,j)\in\mathcal{M}}\mathrm{IoU}(b_i,b_j^{\star})\Bigg)
    \cdot \frac{\min(n_{\mathrm{pred}}, n_{\mathrm{ref}})}{\max(n_{\mathrm{pred}}, n_{\mathrm{ref}})} .
\end{equation}
For pointing, a single predicted point $p$ is scored by a sharpened normalized distance,
\begin{equation}
\label{eq:point}
R = \max\!\Big(0,\; 1 - \frac{\lVert p - p^{\star}\rVert_2}{\delta}\Big),
\qquad \delta = 200,
\end{equation}
while multi-point localization (e.g., segmentation-style contact regions) uses a Chamfer-based reward on image coordinates normalized to $[0,1]$,
\begin{equation}
\label{eq:chamfer}
R = \exp\!\Big(-\frac{d_{\mathrm{CD}}(P, P^{\star})}{\tau_c}\Big),
\qquad
d_{\mathrm{CD}}(P,P^{\star}) = \frac{1}{|P|}\sum_{p\in P}\min_{q\in P^{\star}}\lVert p - q\rVert_2
+ \frac{1}{|P^{\star}|}\sum_{q\in P^{\star}}\min_{p\in P}\lVert q - p\rVert_2,
\end{equation}
with temperature $\tau_c=0.05$.

\paragraph{Trajectories and 3D traces.} For 2D trajectory and path prediction we use a similarity reward based on the (length-normalized) discrete Fr\'echet distance $d_{F}$,
\begin{equation}
\label{eq:traj}
R = \max\!\big(0,\; 1 - \kappa\, d_{F}(y, y^{\star})\big), \qquad \kappa = 2,
\end{equation}
which rewards curves that follow the reference path while penalizing gross deviations. For depth-aware 3D traces we combine a 2D path term with a per-waypoint metric-depth term,
\begin{equation}
\label{eq:trace3d}
R = 0.7\, R_{\mathrm{traj}} + 0.3\, R_{\mathrm{depth}},
\qquad
R_{\mathrm{depth}} = \frac{1}{N}\sum_{k=1}^{N}
\max\!\Big(0,\; 1 - 2\,\frac{\lvert d_k - d_k^{\star}\rvert}{\lvert d_k^{\star}\rvert + \varepsilon}\Big),
\end{equation}
so that trajectory correctness is never masked out by missing or noisy depth predictions.

\paragraph{Discrete and constrained outputs.} For outputs with discrete or strongly constrained formats we use lighter-weight exact-match-style rewards. Multiple-choice, binary judgment, and counting use $R = \mathbb{1}[\hat{y} = y^{\star}]$. Sequential-but-discrete targets such as ordering use a partial-credit reward based on the normalized longest common subsequence,
$R = \mathrm{LCS}(\hat{y}, y^{\star}) / |y^{\star}|$. Continuous estimation tasks (object size, absolute distance, room size) use a regression reward that decays smoothly with relative error,
\begin{equation}
\label{eq:regression}
R = \max\!\Big(0,\; 1 - 2\,\frac{\lvert p - p^{\star}\rvert}{\lvert p^{\star}\rvert + \varepsilon}\Big),
\end{equation}
which provides more informative gradients than hard-threshold accuracy. For direction- and navigation-style spatial-relation questions, we extract the model's \emph{final committed} direction set $\mathcal{D}$ (ignoring intermediate directions mentioned during reasoning) and score it against the reference set $\mathcal{D}^{\star}$ under a ``hit-and-no-contradiction''
rule: $R=1$ if $\mathcal{D}\cap\mathcal{D}^{\star}\neq\varnothing$ and $\mathcal{D}$ contains no axis-opposite of any reference direction, and $R=0$ otherwise.

\paragraph{Open-ended fallback and masking.} For open-ended embodied reasoning whose correctness cannot be robustly determined by rules, we use an LLM-based judge as a fallback,
\begin{equation}
\label{eq:judge}
r_{\mathrm{free}} = J(q, y, y^{\star}) \in \{0,1\}.
\end{equation}
The judge is also used to extract structured answers (choice letters, numbers, directions) when
rule-based parsing fails, so that a correct answer is not penalized for formatting alone. When
the judge service itself fails to return a verdict, we emit a sentinel reward and \emph{mask}
the corresponding sample from the gradient update, ensuring that infrastructure noise is never
confused with an incorrect answer. Overall, this design follows one rule---reward structure
should match output structure---which we find essential for stabilizing RL across heterogeneous
embodied capabilities.

\subsubsection{Stage I: Reasoning-Elicitation RL}
\label{sec:rl-elicit}

The first RL stage acts directly on the cold-started policy $\theta_{\mathrm{sft}}$, whose reasoning is shallow and inconsistent, and rewards trajectories that arrive at verifiably correct action-relevant conclusions across the taxonomy. The RL data is refreshed adaptively around the current model, retaining samples near the capability frontier and keeping the mix
balanced across the taxonomy so that no capability is over-optimized. Because the reward is verifiable and remains dense wherever partial correctness is meaningful, this stage transforms the seeded thinking format into genuine embodied agentic reasoning, producing the policy $\theta_{\mathrm{rl}}$. Rather than training on the RL trajectories directly, which entangle correct and spurious reasoning and reflect a narrow end-of-RL distribution, we use $\theta_{\mathrm{rl}}$ as a \emph{generator} for the rejection-sampling stage of Section~\ref{sec:rft}.

\subsubsection{Stage II: Reward-Specialized RL}
\label{sec:rl-merge}

The second RL stage is applied \emph{after} reasoning has been scaled and consolidated by RFT, starting from the consolidated checkpoint $\theta_{\mathrm{cons}}$ of Section~\ref{sec:rft}. Its purpose is to squeeze out the last increments of precision, and it exploits a structural property of our reward family: the tasks split cleanly into a \emph{continuous-reward} regime and a \emph{discrete-reward} regime, and these two regimes impose conflicting optimization pressures. Continuous rewards, covering pointing, grounding, trajectory, depth, and metric estimation, provide dense, smoothly varying gradients that reward sub-pixel and sub-unit precision and generally favor concise, decisive outputs. Discrete rewards, covering multiple-choice, judgment, ordering, navigation direction, and open-ended reasoning, are sparse and threshold-like, and typically benefit from longer, more exploratory reasoning. Optimizing both with a single policy forces a compromise: the sparse-reward tasks pull the model toward longer chains that slightly blur the fine-grained precision demanded by the dense tasks, and vice versa.

To avoid this interference, we train two specialists from the shared initialization $\theta_{\mathrm{cons}}$ using the same GRPO objective (Eq.~\ref{eq:grpo}) and global batch reward normalization (Eq.~\ref{eq:global-adv}), but on disjoint reward regimes: $\theta_{\mathrm{cont}}$ is optimized on the continuous-reward mixture to maximize geometric and metric precision, and $\theta_{\mathrm{disc}}$ is optimized on the discrete-reward mixture to maximize decision, planning, and reflection accuracy. We then fuse the two sets of parameters into a single model. The fused model $\theta_{\mathrm{final}}$ is the deployed Hy-Embodied-VLM-1.0 A3B: an efficient physical-world agent that unifies precise action-relevant state understanding, grounded action and transition reasoning, and robust sequential and adaptive reasoning within a single $3$B-active-parameter model.

\subsection{Rejection-Sampling Fine-Tuning}
\label{sec:rft}

Rejection-sampling fine-tuning bridges the two RL stages: it turns the reasoning \emph{elicited} by Stage-I RL into a large, clean, and broadly-covering corpus, and then \emph{consolidates} it into a fresh model that becomes the initialization for Stage-II RL. Following the self-evolving paradigm of rejection-sampling fine-tuning \citep{deepseek2025r1}, this proceeds in two steps: synthesizing high-quality long-CoT data with the RL policy (Section~\ref{sec:rollout}), and re-training from the clean pre-training checkpoint (Section~\ref{sec:consolidation}).

\subsubsection{Reasoning Data Synthesis via Rejection Sampling}
\label{sec:rollout}

RL sharpens reasoning but is an expensive way to \emph{scale} it. We therefore use the reasoning-elicited policy $\theta_{\mathrm{rl}}$ (Section~\ref{sec:rl-elicit}) as a generator of high-quality long-CoT data. Concretely, we take queries from the large direct-answer pool of Section~\ref{sec:sft} and, for each query, draw a small number of candidate reasoning trajectories under a best-of-$N$ rollout with $N=4$. Each candidate is then filtered along three complementary axes:
\begin{enumerate}
    \item \textbf{Answer correctness.} The final answer must pass the same verifiable reward
    $R_{\tau}$ used during RL (Section~\ref{sec:reward}), i.e., $R_{\tau}(y,y^{\star})\geq
    \eta_{\tau}$ for a task-specific threshold $\eta_{\tau}$.
    \item \textbf{Thinking quality.} The reasoning trace is scored by an LLM critic for logical
    coherence, absence of repetition, and grounding of intermediate steps in the visual input;
    low-quality or shortcut chains are rejected.
    \item \textbf{Self-consistency.} We favor queries whose retained trajectories agree on the
    final answer across the $N$ samples, using inter-sample agreement as a proxy for the
    reliability of the reasoning rather than a lucky guess.
\end{enumerate}
Because the source queries are drawn from the taxonomy-balanced direct-answer corpus, the synthesized long-CoT data inherits the same broad capability coverage while upgrading the supervision from a bare answer to a verified reasoning trajectory. After filtering, this step yields approximately $1$M high-quality long-CoT examples, which we denote $\mathcal{D}_{\mathrm{cot}}$.

\subsubsection{Reasoning Internalization}
\label{sec:consolidation}

We then internalize the elicited reasoning into a fresh model. Rather than continuing to fine-tune the RL policy, which would carry over its narrow end-of-RL distribution, we restart from the clean pre-training checkpoint $\theta_{\mathrm{pt}}$ and fine-tune on the synthesized long-CoT data together with the remaining direct-answer data. Learning reasoning and direct-answer behaviors jointly from a clean base lets the model absorb long-horizon agentic reasoning while retaining broad, decisive competence. We run this self-evolving cycle for a single iteration, which we find sufficient to internalize stable embodied reasoning, yielding the checkpoint $\theta_{\mathrm{cons}}$ that initializes the reward-specialized RL of Section~\ref{sec:rl-merge}.

\section{Evaluation}
\label{sec:eval}

We evaluate Hy-Embodied-VLM-1.0 A3B as an \emph{efficient physical-world agent} through both capability-oriented benchmarking and closed-loop deployment. Rather than treating embodied intelligence as a collection of isolated visual question-answering tasks, we organize 38 diagnostic benchmarks around three stages of a perception--action loop: understanding the current physical state, reasoning about actions and their state transitions, and maintaining adaptive behavior over extended horizons. We then examine whether these capabilities can be integrated into persistent agent behavior by deploying the model in two complementary vision-language navigation settings.

\paragrapha{Comparison Protocol.}
To disentangle capability improvements from gains obtained simply by increasing model size, our primary ranking on the 38 diagnostic benchmarks compares Hy-Embodied-VLM-1.0 A3B only with models in a comparable active-parameter regime: Hy-Embodied-0.5 MoT-2B~\citep{team2026hy-emb-05}, Qwen3.6-A3B~\citep{qwen36_35b_a3b}, Embodied-R1.5~\citep{yuan2026embodied-r15}, and Cosmos3-Nano~\citep{agarwal2026cosmos3}. Specifically, Embodied-R1.5 has 8B parameters, while the VLM component of Cosmos3-Nano is also 8B, placing both within the scale range considered in our comparison. Hy-Embodied-0.5 A30B~\citep{team2026hy-emb-05} is retained solely as a larger-scale reference and is excluded from the best/second-best ranking and highlighting. To ensure a consistent comparison, we independently evaluate Qwen3.6-A3B, Embodied-R1.5, and Cosmos3-Nano using the same evaluation pipeline; hence, their scores may differ from those reported in the corresponding papers. We also account for differences in the inference modes supported by each model. Embodied-R1.5 is available only in its Instruct configuration and does not provide a separate thinking mode. For Cosmos3-Nano, we observe that enabling thinking leads to substantially degraded performance; we therefore report its non-thinking results. Qwen3.6-A3B and all Hy-Embodied variants, including Hy-Embodied-VLM-1.0 A3B, are evaluated in thinking mode. These configurations are explicitly indicated in the table headers. For each benchmark, we follow its official evaluation protocol, orient the reported metric such that higher is better, and present scores in percentage form. Category averages are computed within each capability category, while the overall average covers all 38 diagnostic benchmarks. The downstream VLN experiments follow task-specific agent formulations, baselines, and metrics, which are described separately in Section~\ref{vln}.

\paragrapha{Overall Results.}
As summarized in Figure~\ref{fig:1-teaser}, Hy-Embodied-VLM-1.0 A3B achieves the highest average score in all three diagnostic capability categories under the parameter-comparable protocol. Across the 38 benchmarks, it ranks first on 19 and second on 11, placing among the top two on 30 benchmarks. Its overall average reaches 65.6, outperforming Qwen3.6-A3B, the strongest comparable baseline overall, by 4.4 points (65.6 vs.\ 61.2) and improving over Hy-Embodied-0.5 MoT-2B by 8.4 points (65.6 vs.\ 57.2). This consistent advantage across state understanding, action--transition reasoning, and adaptive planning shows that the gains cannot be attributed merely to a larger active model. Instead, Hy-Embodied-VLM-1.0 A3B combines broad physical-world competence with a compact active-parameter budget, establishing it as an \textbf{efficient physical-world agent} for practical embodied deployment. Complementing these diagnostic results, the closed-loop evaluations in Section~\ref{vln} show that the same compact model attains leading navigation performance under both route-instruction and object-goal specifications, providing system-level evidence that these capabilities transfer to persistent embodied interaction.

To identify where these gains arise and how they translate into agent behavior, we first structure the following three subsections around the perception--action loop that underpins effective physical-world agency. We begin with how the agent constructs an action-relevant representation of its current environment, then study how it converts that representation into grounded actions and predicts their local effects, and finally evaluate how it maintains progress, plans over extended horizons, and recovers from execution failures. Section~\ref{vln} then moves from capability diagnosis to system-level deployment, testing whether the model can integrate perception, action reasoning, and adaptive planning throughout closed-loop navigation.

\subsection{Action-Relevant State Understanding}
\label{sec:eval_state_understanding}

\begin{table*}[t]
\caption{\textbf{Results on Action-Relevant State Understanding.} Physical and semantic perception, spatial understanding, and robotics-centric understanding. All scores are reported as percentages. The primary comparison is conducted among models with a comparable active-parameter scale; Hy-Embodied-0.5 A30B$^{\dagger}$ is included only as a larger-model reference and is excluded from ranking; Embodied-R1.5 is evaluated in its only available Instruct configuration; Cosmos3-Nano is reported in non-thinking mode because enabling thinking substantially degrades its performance; Qwen3.6-A3B and all Hy-Embodied variants are evaluated in thinking mode. \colorbox{bestcolor}{\hspace{1.2em}} and \colorbox{secondcolor}{\hspace{1.2em}} denote the best and second-best results among the parameter-comparable models, respectively.}
\label{tab:hy_embodied_1_0_cat1}
\centering
\renewcommand{\arraystretch}{1.28}
\setlength{\tabcolsep}{2.6pt}
\setlength{\arrayrulewidth}{0.55pt}
\arrayrulecolor{black}
\normalsize
\begin{threeparttable}
\begin{adjustbox}{max width=\textwidth}
\begin{tabular}{@{}ll|cccccc}
\hline
\multirow{3}{*}{\textbf{Subcategory}} & \multirow{3}{*}{\textbf{Benchmark}} & {\small\bfseries Hy-Embodied} & {\color{black!35}{\small\bfseries Hy-Embodied}} & {\small\bfseries Qwen3.6} & {\small\bfseries Embodied-R1.5} & {\small\bfseries Cosmos3-Nano} & {\small\bfseries Hy-Embodied} \\
 & & {\small\bfseries 0.5 MoT-2B} & {\color{black!35}{\small\bfseries 0.5 A30B}$^{\dagger}$} & {\small\bfseries A3B} & {\small\bfseries 8B} & {\small\bfseries 8B} & {\small\bfseries VLM-1.0 A3B} \\
 & & {\footnotesize thinking} & {\color{black!35}{\footnotesize thinking}} & {\footnotesize thinking} & {\footnotesize Instruct} & {\footnotesize non-thinking} & {\footnotesize thinking} \\
\hline
    \multirow{4}{*}{\shortstack[l]{Physical \& Semantic\\Perception}} & \rule[-1.25ex]{0pt}{4.4ex}BLINK & 82.7 & {\color{black!35}87.1} & \cellcolor{bestcolor}\bfseries 87.9 & 77.8 & 82.4 & \cellcolor{secondcolor} 87.3 \\
     & \rule[-1.25ex]{0pt}{4.4ex}CV-Bench & \cellcolor{secondcolor} 89.2 & {\color{black!35}88.8} & 88.6 & 86.8 & 88.0 & \cellcolor{bestcolor}\bfseries 89.7 \\
     & \rule[-1.25ex]{0pt}{4.4ex}PixMo-Points & 51.4 & {\color{black!35}62.6} & 57.5 & 57.1 & \cellcolor{secondcolor} 59.8 & \cellcolor{bestcolor}\bfseries 64.6 \\
     & \rule[-1.25ex]{0pt}{4.4ex}PointBench & \cellcolor{secondcolor} 69.0 & {\color{black!35}74.8} & 35.1 & 59.1 & 39.2 & \cellcolor{bestcolor}\bfseries 71.7 \\
\hline
    \multirow{12}{*}{\shortstack[l]{Spatial\\Understanding}} & \rule[-1.25ex]{0pt}{4.4ex}Depth-InHouse & 45.7 & {\color{black!35}57.6} & \cellcolor{secondcolor} 63.0 & 52.0 & 47.0 & \cellcolor{bestcolor}\bfseries 67.6 \\
     & \rule[-1.25ex]{0pt}{4.4ex}3DSRBench & \cellcolor{bestcolor}\bfseries 57.0 & {\color{black!35}56.6} & 49.9 & 42.6 & 31.9 & \cellcolor{secondcolor} 52.6 \\
     & \rule[-1.25ex]{0pt}{4.4ex}All-Angles-Bench & 55.1 & {\color{black!35}71.8} & \cellcolor{bestcolor}\bfseries 64.0 & 48.4 & 51.9 & \cellcolor{secondcolor} 63.4 \\
     & \rule[-1.25ex]{0pt}{4.4ex}DA-2K & \cellcolor{bestcolor}\bfseries 92.3 & {\color{black!35}90.2} & 81.4 & 80.5 & 82.8 & \cellcolor{secondcolor} 83.2 \\
     & \rule[-1.25ex]{0pt}{4.4ex}ERQA & 54.5 & {\color{black!35}62.3} & \cellcolor{secondcolor} 57.5 & 37.3 & 45.0 & \cellcolor{bestcolor}\bfseries 60.8 \\
     & \rule[-1.25ex]{0pt}{4.4ex}EmbSpatial-Bench & \cellcolor{secondcolor} 82.8 & {\color{black!35}84.1} & \cellcolor{bestcolor}\bfseries 83.2 & 76.0 & 80.0 & 82.7 \\
     & \rule[-1.25ex]{0pt}{4.4ex}MMSI-Bench & 33.2 & {\color{black!35}39.2} & \cellcolor{bestcolor}\bfseries 41.9 & 29.8 & 34.0 & \cellcolor{secondcolor} 41.8 \\
     & \rule[-1.25ex]{0pt}{4.4ex}MindCube & \cellcolor{secondcolor} 66.3 & {\color{black!35}69.2} & 55.0 & 27.9 & 32.8 & \cellcolor{bestcolor}\bfseries 70.0 \\
     & \rule[-1.25ex]{0pt}{4.4ex}SAT & 76.7 & {\color{black!35}87.3} & \cellcolor{bestcolor}\bfseries 80.7 & 60.7 & 54.0 & \cellcolor{secondcolor} 78.0 \\
     & \rule[-1.25ex]{0pt}{4.4ex}SIBench-mini & 58.2 & {\color{black!35}67.3} & \cellcolor{secondcolor} 60.9 & 51.9 & 52.5 & \cellcolor{bestcolor}\bfseries 64.5 \\
     & \rule[-1.25ex]{0pt}{4.4ex}SITE-Bench-Image & 62.7 & {\color{black!35}74.7} & \cellcolor{secondcolor} 71.7 & 60.3 & 59.6 & \cellcolor{bestcolor}\bfseries 72.3 \\
     & \rule[-1.25ex]{0pt}{4.4ex}ViewSpatial-Bench & \cellcolor{secondcolor} 53.1 & {\color{black!35}59.8} & 49.0 & 43.7 & 52.0 & \cellcolor{bestcolor}\bfseries 53.3 \\
\hline
    \multirow{7}{*}{\shortstack[l]{Robotics-Centric\\Understanding}} & \rule[-1.25ex]{0pt}{4.4ex}OpenEQA & 54.4 & {\color{black!35}73.2} & \cellcolor{bestcolor}\bfseries 73.2 & 53.9 & 53.8 & \cellcolor{secondcolor} 63.1 \\
     & \rule[-1.25ex]{0pt}{4.4ex}PartAfford & 30.1 & {\color{black!35}65.2} & 25.5 & \cellcolor{bestcolor}\bfseries 82.6 & 32.2 & \cellcolor{secondcolor} 63.7 \\
     & \rule[-1.25ex]{0pt}{4.4ex}RoboAfford & \cellcolor{secondcolor} 73.5 & {\color{black!35}75.4} & 66.7 & 60.6 & \cellcolor{bestcolor}\bfseries 76.2 & 71.5 \\
     & \rule[-1.25ex]{0pt}{4.4ex}RoboRefIt & \cellcolor{secondcolor} 82.8 & {\color{black!35}85.2} & 78.5 & 77.2 & 55.4 & \cellcolor{bestcolor}\bfseries 88.2 \\
     & \rule[-1.25ex]{0pt}{4.4ex}RefSpatial-Bench & 45.8 & {\color{black!35}57.2} & \cellcolor{secondcolor} 53.1 & 52.4 & 44.4 & \cellcolor{bestcolor}\bfseries 53.4 \\
     & \rule[-1.25ex]{0pt}{4.4ex}RoboSpatial-Home & 55.7 & {\color{black!35}76.6} & \cellcolor{bestcolor}\bfseries 70.9 & 69.1 & 58.3 & \cellcolor{secondcolor} 69.4 \\
     & \rule[-1.25ex]{0pt}{4.4ex}Where2Place & 68.0 & {\color{black!35}70.0} & 70.0 & \cellcolor{bestcolor}\bfseries 73.0 & \cellcolor{secondcolor} 71.0 & 65.0 \\
\hline
\multicolumn{2}{@{}l|}{\rule[-1.25ex]{0pt}{4.4ex}\textbf{Category Average}} & 62.6 & {\color{black!35}71.1} & \cellcolor{secondcolor} 63.7 & 59.2 & 55.8 & \cellcolor{bestcolor}\bfseries 68.6 \\
\hline
\end{tabular}
\end{adjustbox}
\end{threeparttable}
\end{table*}

\paragrapha{Capability.}
Reliable physical-world action begins with an accurate estimate of the current state. An agent must recognize relevant entities and attributes, recover geometric and viewpoint-dependent relations, and ground objects and affordances into actionable spatial representations. Errors at this stage propagate directly to subsequent decision-making: an agent cannot select a safe interaction or construct a valid plan if it misidentifies the target, its pose, or the surrounding free space. Action-relevant state understanding therefore serves as the perceptual foundation that converts raw visual observations into a structured state on which physical reasoning can operate.

\paragrapha{Benchmarks.}
We evaluate this capability on 23 benchmarks spanning three complementary groups. \emph{Physical and Semantic Perception} assesses object, attribute, and point-level understanding using BLINK~\citep{fu2024blink}, CV-Bench~\citep{tong2024cambrian}, PixMo-Points~\citep{deitke2025molmo}, and PointBench~\citep{cheng2025pointarena}. \emph{Spatial Understanding} evaluates depth, viewpoint, geometry, and spatial-relation reasoning using our internal Depth-InHouse benchmark, 3DSRBench~\citep{ma20253dsrbench}, All-Angles-Bench~\citep{yeh2026seeing}, DA-2K~\citep{yang2024depth}, ERQA~\citep{team2025gemini}, EmbSpatial-Bench~\citep{du2024embspatial}, MMSI-Bench~\citep{yang2025mmsi}, MindCube~\citep{yin2025mindcube}, SAT~\citep{ray2024sat}, SIBench-mini~\citep{yu2025far}, SITE-Bench-Image~\citep{wang2025site}, and ViewSpatial-Bench~\citep{li2025viewspatial}. \emph{Robotics-Centric Understanding} further measures embodied grounding, affordance understanding, and manipulation-relevant scene comprehension using OpenEQA~\citep{majumdar2023openeqa}, PartAfford~\citep{xu2022partafford}, RoboAfford~\citep{tang2025roboafford}, RoboRefIt~\citep{lu2023roborefit}, RefSpatial-Bench~\citep{zhou2025roborefer}, RoboSpatial-Home~\citep{song2025robospatial}, and Where2Place~\citep{yuan2024robopoint}. Together, these benchmarks test whether the model can extract the objects, geometry, spatial relations, and actionable properties required before acting.

\paragrapha{Results.}
As shown in Table~\ref{tab:hy_embodied_1_0_cat1}, Hy-Embodied-VLM-1.0 A3B achieves the highest average among parameter-comparable models, scoring 68.6. This result exceeds Qwen3.6-A3B, the strongest comparable baseline in this category, by 4.9 points (68.6 vs.\ 63.7). Hy-Embodied-VLM-1.0 A3B ranks first on 11 of the 23 benchmarks and second on nine more, placing among the top two on 20 benchmarks. Its leading results span physical perception, fine-grained spatial reasoning, and robotics-centric grounding, rather than concentrating on a single task format. This breadth indicates that the model constructs a robust, action-relevant representation of the physical scene, providing a reliable perceptual foundation for downstream decision-making.

\begin{figure*}[t]
  \centering
  \includegraphics[width=\textwidth]{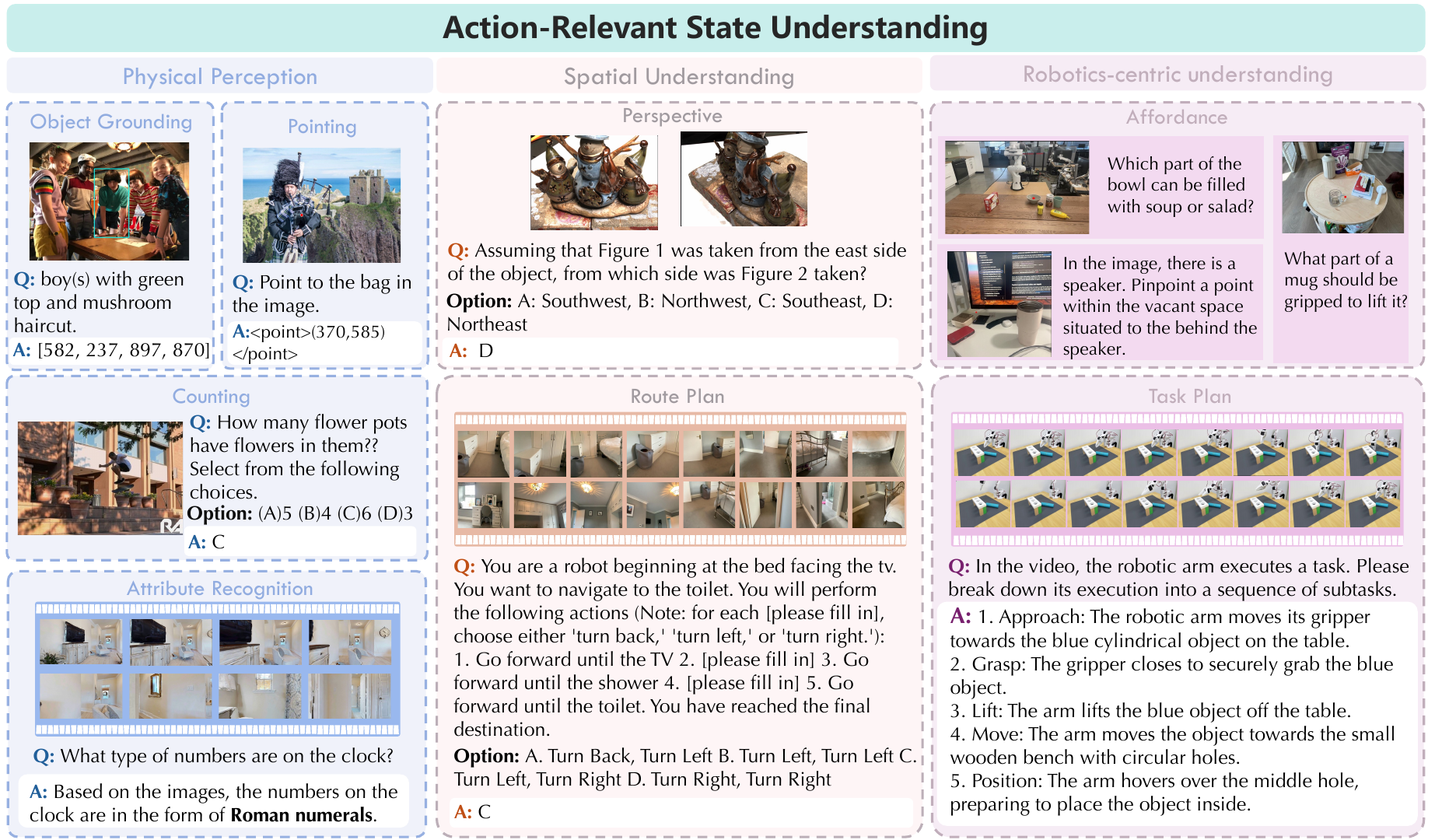}
  \caption{\textbf{Qualitative analysis of action-relevant state understanding.} Hy-Embodied-VLM-1.0 A3B demonstrates physical and semantic perception, spatial understanding, and robotics-centric understanding across counting, pointing, object grounding, attribute recognition, spatial relations, viewpoint reasoning, navigation, affordance grounding, and manipulation-task decomposition.}
  \label{fig:4-perception}
\end{figure*}

\paragrapha{Qualitative Analysis.}
Figure~\ref{fig:4-perception} illustrates how these capabilities appear in representative cases. At the physical and semantic level, Hy-Embodied-VLM-1.0 A3B correctly counts the flower pots containing flowers, localizes a specified bag with a point, grounds the boys matching a compositional attribute description, and recognizes the Roman numerals on a clock. The spatial examples require more than object recognition: the model distinguishes whether a car lies directly beneath a road sign, infers the orientation of a bench relative to the camera, and converts a sequence of egocentric room observations into the correct left--right route decisions. In robotics-centric scenes, it identifies task-specific affordances---such as where a bowl can be filled, where a mug should be grasped, and which free region is suitable for placement---and decomposes a manipulation video into meaningful subtasks. Together, these cases show that the model links semantic perception with geometry and actionability, producing the structured scene understanding needed for physical interaction.

\subsection{Action-Transition Reasoning}
\label{sec:eval_transition_reasoning}

\paragrapha{Capability.}
Recognizing the current state is necessary but insufficient for acting in the physical world. An agent must also infer the intent and dynamics of an interaction, choose an action that is feasible in the present configuration, ground that action to the correct object or direction, and anticipate its immediate physical effect. Action-transition reasoning provides this bridge from perception to control: it maps a state and goal to an executable local decision while accounting for how that decision changes the environment.

\paragrapha{Benchmarks.}
We assess this capability with eight benchmarks organized into three groups. \emph{Interaction Understanding} is evaluated with FineBench~\citep{faure2026finebench} and CrossHOI-Bench~\citep{lei2026crosshoi}, which require recognizing human--object and cross-object interactions. \emph{Action Decision and Grounding} is evaluated with PIO~\citep{xue2025pio}, VABench-Point and VABench-Visual-Trace~\citep{yuan2025vabench}, and the affordance and trajectory tracks of ShareRobot-Bench~\citep{ji2025robobrain}; these tasks test interaction-point localization, affordance prediction, action grounding, and trajectory-level reasoning. Finally, RoboBench-MCQ~\citep{luo2025robobench} evaluates robotics-oriented action selection and local task reasoning. Collectively, these benchmarks measure whether a model can convert an observed physical state into a grounded action while anticipating the resulting local transition.

\begin{table*}[t]
\caption{\textbf{Results on Action--Transition Reasoning.} Interaction understanding, action decision and grounding, and local robotic task reasoning. All scores are reported as percentages. The primary comparison is conducted among models with a comparable active-parameter scale; Hy-Embodied-0.5 A30B$^{\dagger}$ is included only as a larger-model reference and is excluded from ranking; Embodied-R1.5 is evaluated in its only available Instruct configuration; Cosmos3-Nano is reported in non-thinking mode because enabling thinking substantially degrades its performance; Qwen3.6-A3B and all Hy-Embodied variants are evaluated in thinking mode. \colorbox{bestcolor}{\hspace{1.2em}} and \colorbox{secondcolor}{\hspace{1.2em}} denote the best and second-best results among the parameter-comparable models, respectively.}
\label{tab:hy_embodied_1_0_cat2}
\centering
\renewcommand{\arraystretch}{1.28}
\setlength{\tabcolsep}{2.6pt}
\setlength{\arrayrulewidth}{0.55pt}
\arrayrulecolor{black}
\normalsize
\begin{threeparttable}
\begin{adjustbox}{max width=\textwidth}
\begin{tabular}{@{}ll|cccccc}
\hline
\multirow{3}{*}{\textbf{Subcategory}} & \multirow{3}{*}{\textbf{Benchmark}} & {\small\bfseries Hy-Embodied} & {\color{black!35}{\small\bfseries Hy-Embodied}} & {\small\bfseries Qwen3.6} & {\small\bfseries Embodied-R1.5} & {\small\bfseries Cosmos3-Nano} & {\small\bfseries Hy-Embodied} \\
 & & {\small\bfseries 0.5 MoT-2B} & {\color{black!35}{\small\bfseries 0.5 A30B}$^{\dagger}$} & {\small\bfseries A3B} & {\small\bfseries 8B} & {\small\bfseries 8B} & {\small\bfseries VLM-1.0 A3B} \\
 & & {\footnotesize thinking} & {\color{black!35}{\footnotesize thinking}} & {\footnotesize thinking} & {\footnotesize Instruct} & {\footnotesize non-thinking} & {\footnotesize thinking} \\
\hline
    \multirow{2}{*}{\shortstack[l]{Interaction\\Understanding}} & \rule[-1.25ex]{0pt}{4.4ex}FineBench & 56.9 & {\color{black!35}61.4} & \cellcolor{secondcolor} 76.9 & 67.1 & 63.5 & \cellcolor{bestcolor}\bfseries 80.3 \\
     & \rule[-1.25ex]{0pt}{4.4ex}CrossHOI-Bench & 40.7 & {\color{black!35}52.7} & \cellcolor{secondcolor} 58.0 & 55.1 & 51.0 & \cellcolor{bestcolor}\bfseries 63.2 \\
\hline
    \multirow{5}{*}{\shortstack[l]{Action Decision\\\& Grounding}} & \rule[-1.25ex]{0pt}{4.4ex}PIO & 54.6 & {\color{black!35}65.3} & 47.9 & \cellcolor{secondcolor} 61.6 & 54.4 & \cellcolor{bestcolor}\bfseries 65.3 \\
     & \rule[-1.25ex]{0pt}{4.4ex}VABench-Point & 26.0 & {\color{black!35}57.8} & 50.5 & \cellcolor{bestcolor}\bfseries 61.4 & 45.2 & \cellcolor{secondcolor} 59.7 \\
     & \rule[-1.25ex]{0pt}{4.4ex}VABench-Visual-Trace & 75.0 & {\color{black!35}72.0} & 80.3 & \cellcolor{bestcolor}\bfseries 89.8 & \cellcolor{secondcolor} 81.6 & 79.7 \\
     & \rule[-1.25ex]{0pt}{4.4ex}ShareRobot-Bench-Affordance & \cellcolor{secondcolor} 26.8 & {\color{black!35}28.6} & \cellcolor{bestcolor}\bfseries 28.2 & 25.2 & 23.0 & 26.7 \\
     & \rule[-1.25ex]{0pt}{4.4ex}ShareRobot-Bench-Trajectory & \cellcolor{secondcolor} 73.3 & {\color{black!35}76.9} & 68.9 & 69.2 & 65.5 & \cellcolor{bestcolor}\bfseries 76.7 \\
\hline
    \multirow{1}{*}[0.9ex]{\shortstack[l]{Robot Action \&\\Local Reasoning}} & \rule[-1.45ex]{0pt}{5.2ex}RoboBench-MCQ & 49.2 & {\color{black!35}62.8} & \cellcolor{secondcolor} 59.1 & 41.1 & 43.5 & \cellcolor{bestcolor}\bfseries 61.2 \\
\hline
\multicolumn{2}{@{}l|}{\rule[-1.25ex]{0pt}{4.4ex}\textbf{Category Average}} & 50.3 & {\color{black!35}59.7} & 58.7 & \cellcolor{secondcolor} 58.8 & 53.5 & \cellcolor{bestcolor}\bfseries 64.1 \\
\hline
\end{tabular}
\end{adjustbox}
\end{threeparttable}
\end{table*}

\paragrapha{Results.}
Table~\ref{tab:hy_embodied_1_0_cat2} shows the clearest advantage of Hy-Embodied-VLM-1.0 A3B. It achieves an average of 64.1, outperforming Embodied-R1.5, the strongest parameter-comparable baseline in this category, by 5.3 points (64.1 vs.\ 58.8). The model ranks first on five of the eight benchmarks---FineBench, CrossHOI-Bench, PIO, ShareRobot-Bench-Trajectory, and RoboBench-MCQ---and second on VABench-Point. These gains consistently cover interaction recognition, action grounding, trajectory prediction, and local robotic decision-making. The results therefore show that the model does more than perceive a scene accurately: it can convert observations into grounded action choices and reason about the physical transitions induced by those actions. This strong action-oriented reasoning is central to its role as an efficient physical-world agent.

\begin{figure*}[t]
  \centering
  \includegraphics[width=\textwidth]{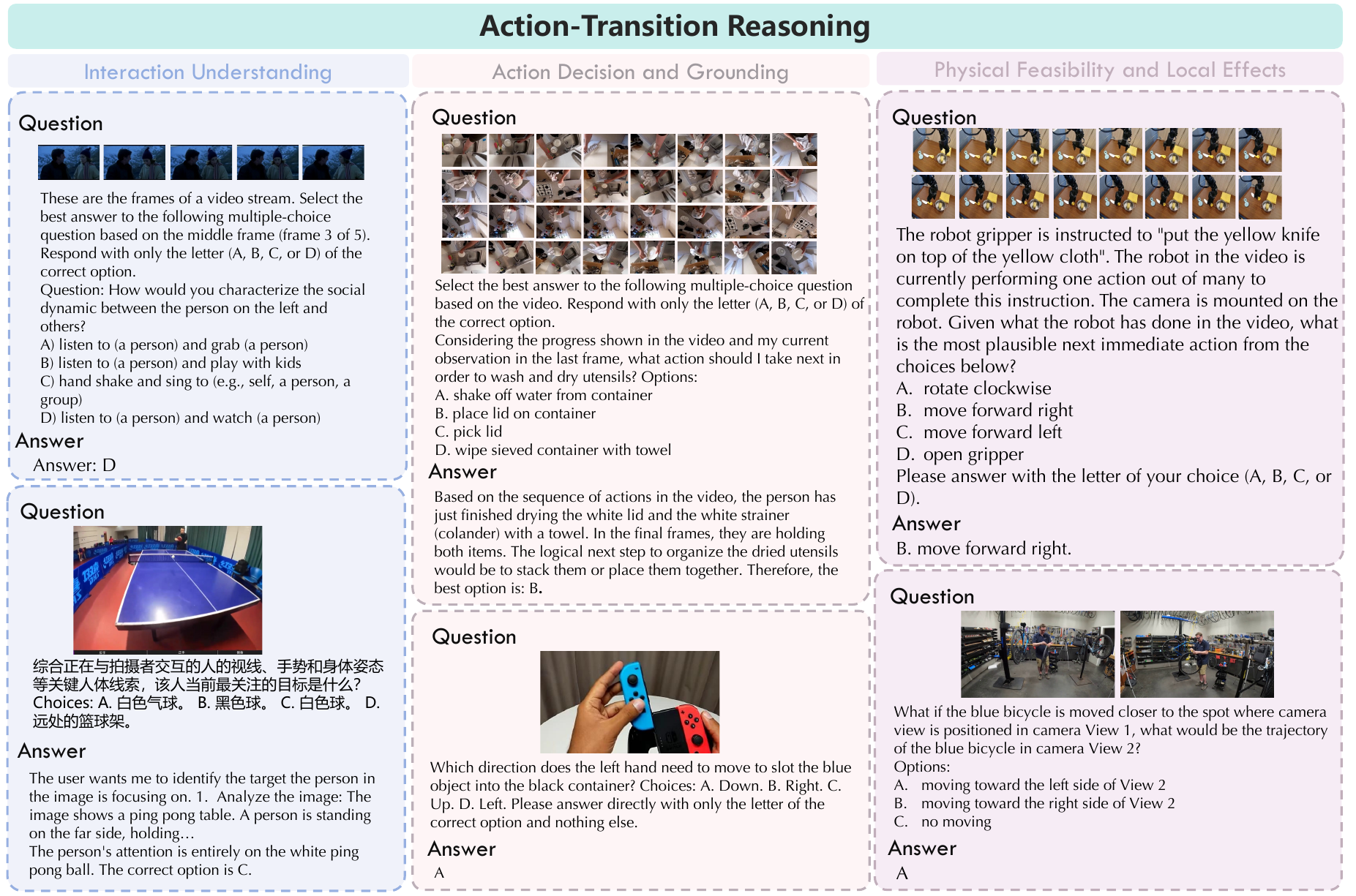}
  \caption{\textbf{Qualitative analysis of action-transition reasoning.} Hy-Embodied-VLM-1.0 A3B recognizes interaction intent, selects progress-consistent actions, grounds directional decisions, and reasons about local physical feasibility and viewpoint-dependent effects.}
  \label{fig:5-reasoning}
\end{figure*}

\paragrapha{Qualitative Analysis.}
The cases in Figure~\ref{fig:5-reasoning} reveal how Hy-Embodied-VLM-1.0 A3B reasons across the state--action boundary. For interaction understanding, it identifies the social dynamics in a short video and determines the object attended to by a person from gaze, pose, and scene context. For action decision and grounding, the model tracks the progress of a utensil-cleaning sequence to select ``place lid on container'' as the appropriate next step, and infers the required downward motion for inserting a controller component into its slot. The physical-feasibility cases further require embodied frame transformation: the model selects a forward-right gripper motion to place a knife on a cloth and predicts how moving a bicycle in one camera view changes its apparent trajectory in another. These examples demonstrate that the model jointly uses interaction semantics, temporal progress, spatial grounding, and local physical constraints to select actions whose effects are consistent with the scene.

\subsection{Sequential and Adaptive Reasoning}
\label{sec:eval_adaptive_reasoning}

\paragrapha{Capability.}
Real-world tasks unfold over multiple steps and rarely proceed exactly as planned. A capable agent must preserve relevant history, determine how far execution has progressed, compose local actions into a long-horizon plan, detect when an intended subtask has failed, and revise its behavior accordingly. Sequential and adaptive reasoning enables persistent agency: it prevents the model from treating each observation independently and allows it to continue making progress under changing or imperfect execution conditions.

\paragrapha{Benchmarks.}
We evaluate this capability on seven benchmarks spanning three groups. \emph{History- and Progress-Aware Reasoning} uses SITE-Bench-Video~\citep{wang2025site}, VSIBench~\citep{yang2025thinking}, EgoPlan2~\citep{qiu2024egoplan}, and Cosmos~\citep{azzolini2025cosmos} to test whether the model can integrate temporal observations with action history and task progress. \emph{Long-Horizon Planning and Composition} is evaluated on VLABench~\citep{zhang2025vlabench} and RoboBench-Planning~\citep{luo2025robobench}, which require composing multiple actions toward a distant objective. Finally, RoboFAC~\citep{ye2025robofac} evaluates \emph{Reflection, Repair, and Recovery}, testing whether the model can identify a failure and revise its behavior. Together, these benchmarks assess the temporal memory, planning, self-correction, and recovery capabilities required in dynamic physical environments.

\begin{table*}[t]
\caption{\textbf{Results on Sequential and Adaptive Reasoning.} History-aware reasoning, long-horizon planning, and reflection and recovery. All scores are reported as percentages. The primary comparison is conducted among models with a comparable active-parameter scale; Hy-Embodied-0.5 A30B$^{\dagger}$ is included only as a larger-model reference and is excluded from ranking; Embodied-R1.5 is evaluated in its only available Instruct configuration; Cosmos3-Nano is reported in non-thinking mode because enabling thinking substantially degrades its performance; Qwen3.6-A3B and all Hy-Embodied variants are evaluated in thinking mode. \colorbox{bestcolor}{\hspace{1.2em}} and \colorbox{secondcolor}{\hspace{1.2em}} denote the best and second-best results among the parameter-comparable models, respectively.}
\label{tab:hy_embodied_1_0_cat3}
\centering
\renewcommand{\arraystretch}{1.28}
\setlength{\tabcolsep}{2.6pt}
\setlength{\arrayrulewidth}{0.55pt}
\arrayrulecolor{black}
\normalsize
\begin{threeparttable}
\begin{adjustbox}{max width=\textwidth}
\begin{tabular}{@{}ll|cccccc}
\hline
\multirow{3}{*}{\textbf{Subcategory}} & \multirow{3}{*}{\textbf{Benchmark}} & {\small\bfseries Hy-Embodied} & {\color{black!35}{\small\bfseries Hy-Embodied}} & {\small\bfseries Qwen3.6} & {\small\bfseries Embodied-R1.5} & {\small\bfseries Cosmos3-Nano} & {\small\bfseries Hy-Embodied} \\
 & & {\small\bfseries 0.5 MoT-2B} & {\color{black!35}{\small\bfseries 0.5 A30B}$^{\dagger}$} & {\small\bfseries A3B} & {\small\bfseries 8B} & {\small\bfseries 8B} & {\small\bfseries VLM-1.0 A3B} \\
 & & {\footnotesize thinking} & {\color{black!35}{\footnotesize thinking}} & {\footnotesize thinking} & {\footnotesize Instruct} & {\footnotesize non-thinking} & {\footnotesize thinking} \\
\hline
    \multirow{4}{*}{\shortstack[l]{History- \& Progress-\\Aware Reasoning}} & \rule[-1.25ex]{0pt}{4.4ex}SITE-Bench-Video & 63.5 & {\color{black!35}72.5} & \cellcolor{bestcolor}\bfseries 71.1 & 59.1 & 57.6 & \cellcolor{secondcolor} 69.2 \\
     & \rule[-1.25ex]{0pt}{4.4ex}VSIBench & \cellcolor{bestcolor}\bfseries 60.5 & {\color{black!35}68.3} & 57.5 & \cellcolor{secondcolor} 59.2 & 50.4 & 58.9 \\
     & \rule[-1.25ex]{0pt}{4.4ex}EgoPlan2 & 45.5 & {\color{black!35}51.4} & \cellcolor{secondcolor} 49.9 & \cellcolor{bestcolor}\bfseries 61.0 & 42.6 & 49.6 \\
     & \rule[-1.25ex]{0pt}{4.4ex}Cosmos & 54.3 & {\color{black!35}65.5} & \cellcolor{secondcolor} 67.8 & \cellcolor{bestcolor}\bfseries 68.6 & 67.1 & 66.9 \\
\hline
    \multirow{2}{*}{\shortstack[l]{Long-Horizon Planning\\\& Composition}} & \rule[-1.25ex]{0pt}{4.4ex}VLABench & 16.2 & {\color{black!35}51.0} & \cellcolor{secondcolor} 49.9 & 39.4 & 48.9 & \cellcolor{bestcolor}\bfseries 51.1 \\
     & \rule[-1.25ex]{0pt}{4.4ex}RoboBench-Planning & \cellcolor{secondcolor} 54.2 & {\color{black!35}59.3} & 53.9 & 39.4 & 41.5 & \cellcolor{bestcolor}\bfseries 54.9 \\
\hline
    \multirow{1}{*}[1.35ex]{\shortstack[l]{Reflection, Repair\\\& Recovery}} & \rule[-1.45ex]{0pt}{5.6ex}RoboFAC & 35.6 & {\color{black!35}42.8} & 41.4 & \cellcolor{secondcolor} 43.9 & 34.4 & \cellcolor{bestcolor}\bfseries 51.0 \\
\hline
\multicolumn{2}{@{}l|}{\rule[-1.25ex]{0pt}{4.4ex}\textbf{Category Average}} & 47.1 & {\color{black!35}58.7} & \cellcolor{secondcolor} 55.9 & 52.9 & 48.9 & \cellcolor{bestcolor}\bfseries 57.4 \\
\hline
\end{tabular}
\end{adjustbox}
\end{threeparttable}
\end{table*}

\paragrapha{Results.}
As reported in Table~\ref{tab:hy_embodied_1_0_cat3}, Hy-Embodied-VLM-1.0 A3B again achieves the highest average among parameter-comparable models, reaching 57.4 and outperforming Qwen3.6-A3B by 1.4 points (57.4 vs.\ 55.9). It ranks first on VLABench, RoboBench-Planning, and RoboFAC and second on SITE-Bench-Video. The model consequently demonstrates competitive capability across temporal understanding, long-horizon task composition, reflection, and failure recovery---abilities that are particularly challenging for compact models but essential for persistent interaction in dynamic environments.

\begin{figure*}[t]
  \centering
  \includegraphics[width=\textwidth]{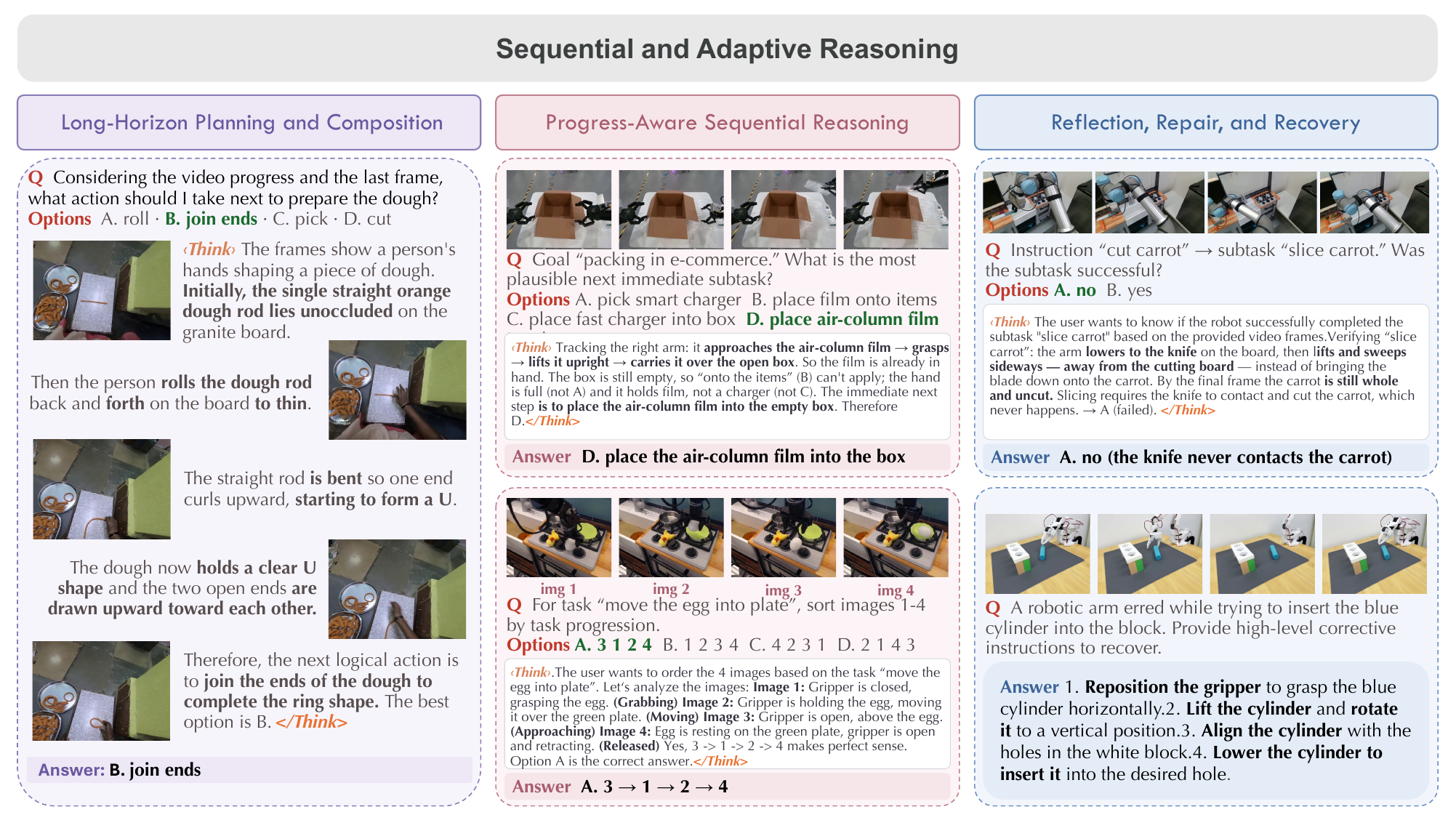}
  \caption{\textbf{Qualitative analysis of sequential and adaptive reasoning.} Hy-Embodied-VLM-1.0 A3B performs long-horizon action composition, progress-aware temporal reasoning, failure diagnosis, and corrective recovery across human and robotic task sequences.}
  \label{fig:6-sequential}
\end{figure*}

\paragrapha{Qualitative Analysis.}
Figure~\ref{fig:6-sequential} demonstrates three forms of reasoning required for sustained task execution. In long-horizon planning, the model traces how a straight dough rod is progressively rolled and bent into a U shape, then identifies joining the two ends as the next action needed to complete the ring. In progress-aware reasoning, it uses the current contents of a package to select placement of the air-column film as the next subtask and correctly orders shuffled frames of an egg-transfer sequence from approaching, to grasping, transporting, and finally releasing the egg. The reflection examples test whether the model can distinguish intended from actual outcomes: it recognizes that a carrot-cutting attempt failed because the knife never contacted the carrot, and, after a failed cylinder insertion, proposes a concrete recovery plan that reorients the grasp, aligns the cylinder with the target hole, and retries insertion. Rather than merely producing a plausible next action, the model explicitly tracks state changes, diagnoses execution failures, and adapts the plan to recover.

\subsection{Vision-Language Navigation}
\label{vln}

\begin{figure*}[t]
  \centering
  \includegraphics[width=\textwidth]{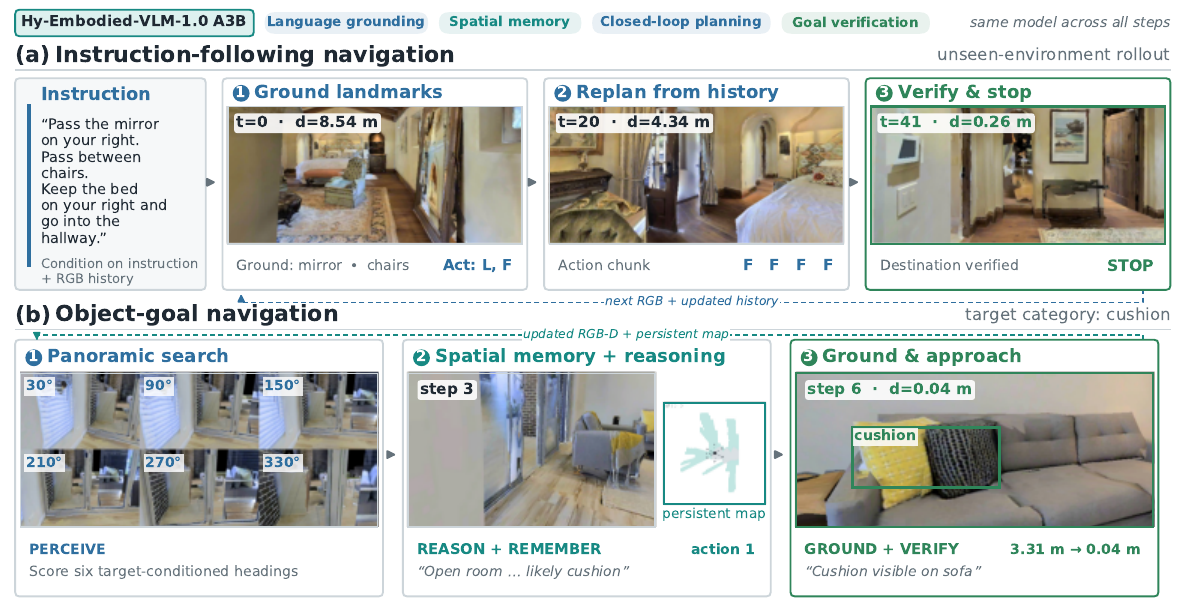}
  \caption{\textbf{Closed-loop navigation with Hy-Embodied-VLM-1.0 A3B.}
  R2R-CE and MP3D ObjectNav trajectories illustrate how the policy maps successive observations and navigation states to action sequences, replanning decisions, subgoals, and termination predictions.}
  \label{fig:navigation-agent-demo}
\end{figure*}

We deploy Hy-Embodied-VLM-1.0 A3B as the decision-making core of two closed-loop navigation agents to evaluate persistent embodied behavior.
In both settings, the model operates within a recurrent loop of perception and action: each environment transition updates the observation history and navigation state on which subsequent decisions are conditioned.
R2R-CE~\citep{anderson2018vision,krantz_vlnce_2020} evaluates language-conditioned trajectory execution, whereas zero-shot Object Goal Navigation on Matterport3D (MP3D)~\citep{chang2017matterport3d} evaluates category-conditioned exploration without route-level guidance.
Together, they provide a system-level evaluation of agentic, long-horizon decision-making under two complementary goal specifications.
Representative closed-loop rollouts of the resulting navigation agents are illustrated in Figure~\ref{fig:navigation-agent-demo}.

\subsubsection{Instruction-Following Navigation on R2R-CE}

\paragrapha{Agent Formulation and Evaluation.}
R2R-CE instantiates Room-to-Room navigation in continuous Matterport3D environments using Habitat~\citep{savva2019habitat}.
We evaluate on the standard \texttt{val-unseen} split, whose environments are disjoint from training.
At each decision step, the policy conditions on the route instruction and accumulated egocentric RGB history.
Using role-specific prompts with shared model parameters, the policy performs both action prediction and termination assessment.
The model outputs either \texttt{stop} or a finite sequence of low-level navigation actions.
Following execution, the newly acquired observation is appended to the history before the policy is queried again.
This receding-horizon protocol evaluates instruction grounding, temporal state tracking, corrective decision-making, and terminal-state recognition over extended trajectories.
We report Success Rate (SR), Oracle Success (OS), Success weighted by Path Length (SPL), and Navigation Error (NE) under the standard 3\,m success criterion.

\newsavebox{\navRRcebox}
\savebox{\navRRcebox}{%
  \footnotesize\setlength{\tabcolsep}{5pt}%
  \begin{tabular}{lcccc}
    \toprule
    Model & SR $\uparrow$ & OS $\uparrow$ & SPL $\uparrow$ & NE $\downarrow$ \\
    \midrule
    Uni-NaVid~\citeyearpar{zhang2024uninavid} & 47.0 & 53.3 & 42.7 & 5.5 \\
    InternVLA-N1~\citeyearpar{wei2026dualvln} & 55.4 & 60.6 & \cellcolor{secondcolor} 52.1 & \cellcolor{secondcolor} 4.8 \\
    Vesta~\citeyearpar{bjorck2026vesta} & 55.5 & 61.4 & 50.8 & 5.1 \\
    Qwen-VLA-Base~\citeyearpar{wang2026qwenvla} & 53.8 & 61.7 & 49.4 & 5.2 \\
    Qwen-VLA-Instruct~\citeyearpar{wang2026qwenvla} & \cellcolor{secondcolor} 57.5 & \cellcolor{bestcolor}\bfseries 69.0 & 51.2 & 5.1 \\
    \midrule
    \textbf{Hy-Embodied-VLM-1.0 A3B} & \cellcolor{bestcolor}\bfseries 57.9 & \cellcolor{secondcolor} 66.4 & \cellcolor{bestcolor}\bfseries 54.2 & \cellcolor{bestcolor}\bfseries 4.5 \\
    \bottomrule
  \end{tabular}%
}
\newsavebox{\navObjbox}
\savebox{\navObjbox}{%
  \footnotesize\setlength{\tabcolsep}{6pt}%
  \begin{tabular}{lcc}
    \toprule
    Model & SR $\uparrow$ & SPL $\uparrow$ \\
    \midrule
    Qwen3.5-35B-A3B~\citeyearpar{qwen35} & 35.8 & 9.7 \\
    Qwen3.6-35B-A3B~\citeyearpar{qwen36_35b_a3b} & \cellcolor{secondcolor} 36.1 & \cellcolor{secondcolor} 9.8 \\
    \midrule
    \textbf{Hy-Embodied-VLM-1.0 A3B} & \cellcolor{bestcolor}\bfseries 38.3 & \cellcolor{bestcolor}\bfseries 11.2 \\
    \bottomrule
  \end{tabular}%
}

\begin{table}[t]
  \centering
  \caption{\textbf{Closed-loop navigation results.}
  Subtable (a) reports map-free, RGB-only R2R-CE \texttt{val-unseen}. Subtable (b) reports zero-shot MP3D Object Goal Navigation with RGB-D input and identical geometric processing and spatial-memory modules across models.
  SR, OS, and SPL are percentages and NE is in meters. \colorbox{bestcolor}{\hspace{1.2em}} and \colorbox{secondcolor}{\hspace{1.2em}} denote the best and second-best results, respectively.}
  \label{tab:closed_loop_navigation}
  \begin{adjustbox}{max width=\linewidth,center}
    \begin{subtable}[t]{\wd\navRRcebox}
      \centering
      \caption{R2R-CE \texttt{val-unseen} (RGB-only, map-free).}
      \label{tab:r2r_vln_results}
      \usebox{\navRRcebox}
    \end{subtable}%
    \hspace{1.5em}%
    \begin{subtable}[t]{\wd\navObjbox}
      \centering
      \caption{MP3D Object Goal Navigation (RGB-D, zero-shot).}
      \label{tab:mp3d_objectnav_results}
      \begin{minipage}[c][\dimexpr\ht\navRRcebox+\dp\navRRcebox\relax][c]{\linewidth}
        \centering
        \usebox{\navObjbox}
      \end{minipage}
    \end{subtable}%
  \end{adjustbox}
\end{table}

\paragrapha{Results.}
As shown in Table~\ref{tab:r2r_vln_results}, Hy-Embodied-VLM-1.0 A3B achieves the best SR (57.9\%), SPL (54.2\%), and NE (4.5\,m), while ranking second in OS (66.4\%).
Compared with the strongest baseline for each metric, it improves SR and SPL by 0.4 and 2.1 percentage points, respectively, and reduces NE by 0.3\,m.
Although Qwen-VLA-Instruct obtains a higher OS, Hy-Embodied-VLM-1.0 A3B achieves higher SR with lower NE, a pattern consistent with more reliable terminal-state estimation once the trajectory approaches the destination.
Its leading SPL additionally demonstrates higher success-weighted path efficiency during closed-loop execution.

\subsubsection{Zero-Shot Object-Goal Navigation on MP3D}

\paragrapha{Evaluation Protocol.}
Object Goal Navigation removes route-level guidance: given only a target category and RGB-D observations, the policy must locate a corresponding instance in an unseen environment and terminate within its success region.
The navigation framework derives traversability candidates from depth and maintains an explicit spatial representation across successive observations.
Within this framework, Hy-Embodied-VLM-1.0 A3B performs direction scoring, subgoal selection, target grounding, and action selection.
Following each environment transition, the estimated pose and newly acquired observation update the spatial state before the next policy query.
All models are evaluated with identical geometric processing and spatial-memory modules, and no model receives ObjectNav-specific parameter adaptation.
This controlled protocol evaluates model-level differences in semantic exploration and sequential decision-making.
We report SR and SPL.

\paragrapha{Results.}
As shown in Table~\ref{tab:mp3d_objectnav_results}, Hy-Embodied-VLM-1.0 A3B achieves an SR of 38.3\% and an SPL of 11.2\%.
It outperforms the stronger Qwen3.6-35B-A3B baseline by 2.2 percentage points in SR and 1.4 points in SPL.
Because Object Goal Navigation requires route construction through exploration rather than execution of a provided route description, these gains provide complementary evidence for category-conditioned planning, persistent search, and efficient termination.

\paragrapha{Qualitative Agent Rollouts.}
The complete trajectories in Figs.~\ref{fig:r2r-ce-qualitative} and~\ref{fig:mp3d-objectnav-qualitative} complement the aggregate metrics by visualizing the evolution of observations and decisions over an episode.
They illustrate instruction following and semantic exploration as temporally extended closed-loop processes rather than isolated action predictions.


\section{Conclusion}

In this report, we introduce Hy-Embodied-VLM-1.0, an efficient embodied foundation model for physical-world understanding, reasoning, and decision-making. We develop an action-centric capability taxonomy spanning action-relevant state understanding, action-transition reasoning, and sequential and adaptive reasoning, and use it to guide data construction, training, and evaluation. Built on the Hy3-A3B backbone and Hy-ViT2 vision encoder, the model activates only approximately 3B parameters per token while maintaining strong embodied capabilities. Across 38 benchmarks, Hy-Embodied-VLM-1.0 achieves the best overall performance among models of comparable scale and improves the average score of Hy-Embodied-0.5 MoT-2B by 8.4\%. It also demonstrates strong performance on vision-language navigation tasks requiring multi-step interaction and long-horizon reasoning. We hope that Hy-Embodied-VLM-1.0 provides an efficient and practical foundation for future physical-world agents capable of understanding states, reasoning about actions and transitions, and adapting over time.

\newpage

\renewcommand{\refname}{References}
\renewcommand{\bibname}{References}
\renewcommand{\bibsection}{\section*{\raggedright \Large References}}
\makeatother
\bibliographystyle{abbrvnat}
\bibliography{egbib}

\newpage

\appendix
\section{Contributors}

\vspace{1em}
\begin{itemize}[label=$\bullet$, leftmargin=*, itemsep=0.8em]
\item \textbf{Project Supervisors: } Han Hu, Zhengyou Zhang
\item \textbf{Project Leader: } Yongming Rao
\item \textbf{Core Contributors: } Ziyi Wang, Xumin Yu, Yonggen Ling
\item \textbf{Contributors:  } Yunheng Li, Oran Wang, Mingqi Gao, Yuchen Zhou, Yves Liang, Zuyan Liu, Yani Zhang, Rui Huang, Xiaoran Xu, Bowen Yuan, Yifu Yuan, Xu Tan, He Zhang, Yufei Huang, Shenghao Zhang, Hongsheng Wu
\end{itemize}

\newpage
\section{Visualizations}

\newcounter{appendixfigure}
\renewcommand{\theappendixfigure}{B\arabic{appendixfigure}}

\noindent
\textbf{Physical Perception.}

\begin{center}
    \includegraphics[width=0.92\textwidth]{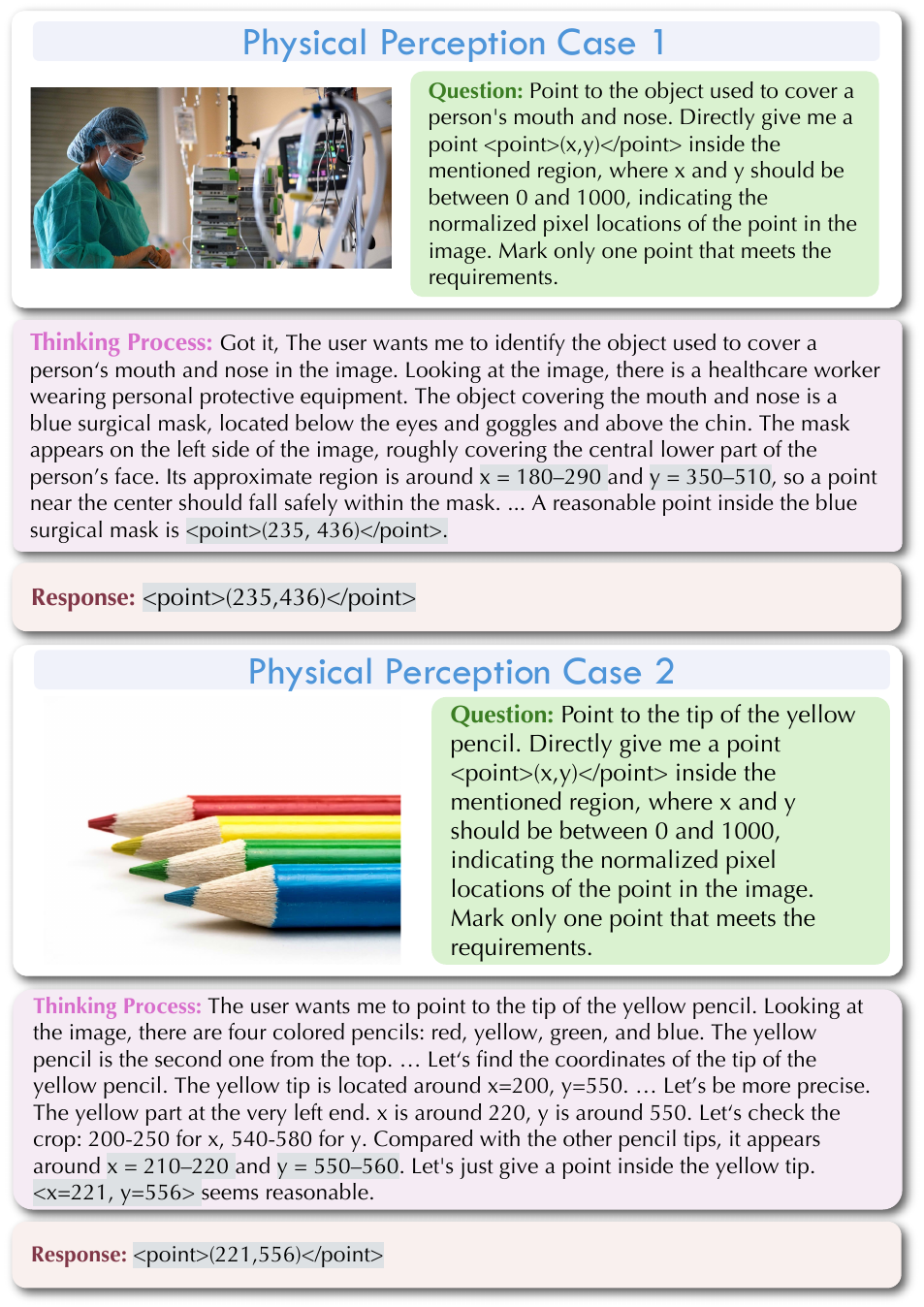}

    \vspace{0.5em}
    \refstepcounter{appendixfigure}
    \small Fig.\theappendixfigure: \textbf{Examples of physical perception.}
\end{center}

\clearpage
\noindent
\textbf{Spatial Understanding Case 1.}

\begin{center}
    \includegraphics[width=0.92\textwidth]{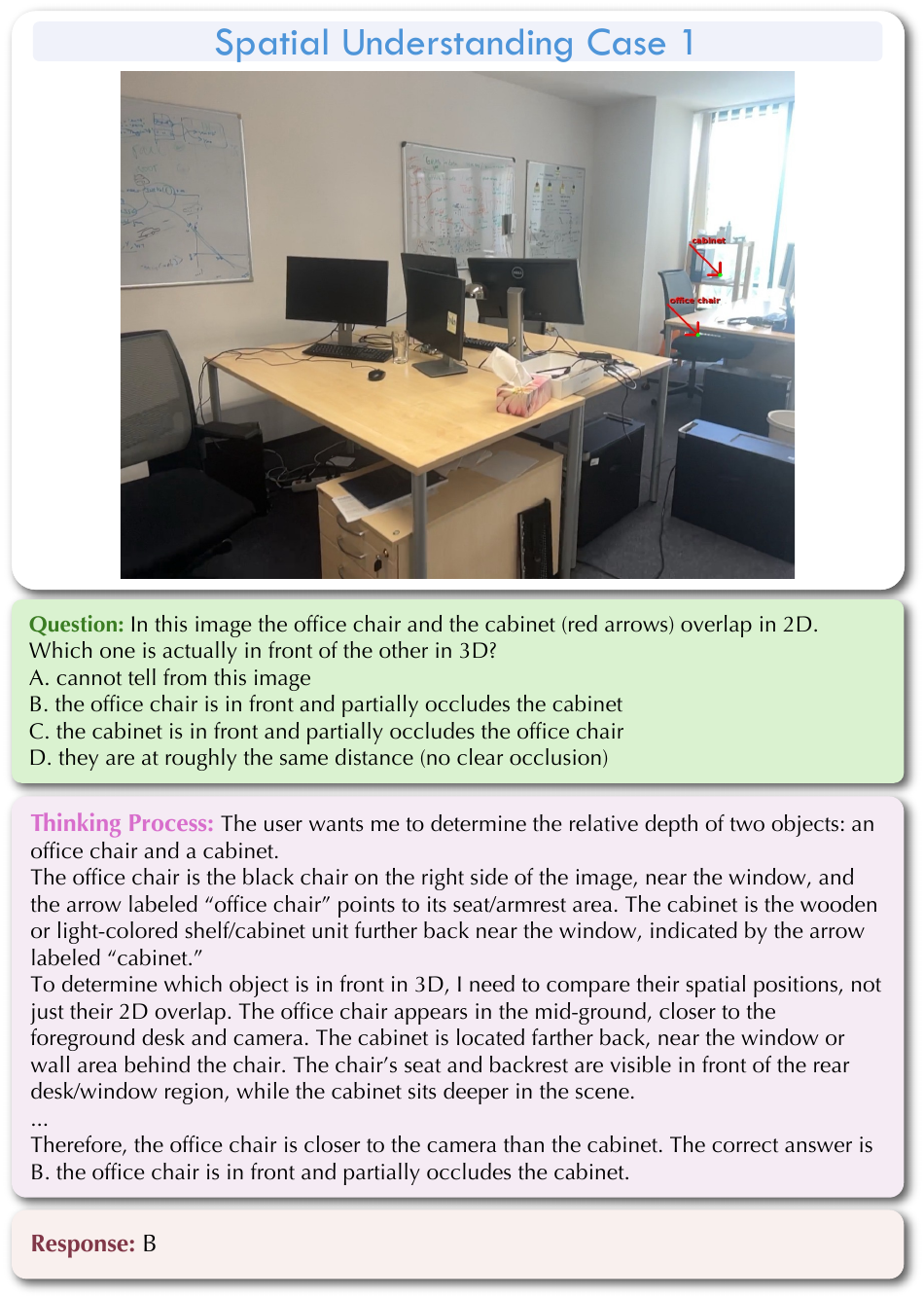}

    \vspace{0.5em}
    \refstepcounter{appendixfigure}
    \small Fig.\theappendixfigure: \textbf{Example of spatial understanding, case 1.}
\end{center}

\clearpage
\noindent
\textbf{Spatial Understanding Case 2.}

\begin{center}
    \includegraphics[width=0.92\textwidth]{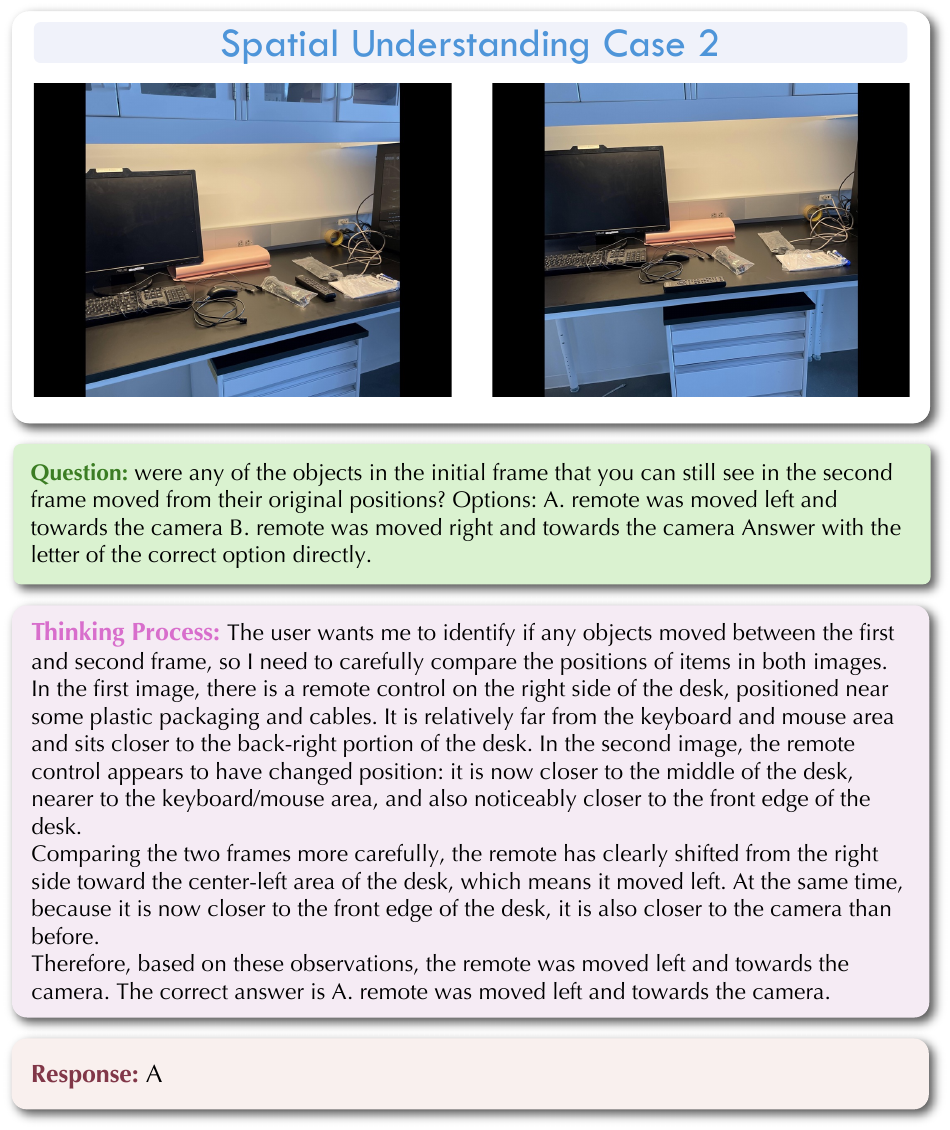}

    \vspace{0.5em}
    \refstepcounter{appendixfigure}
    \small Fig.\theappendixfigure: \textbf{Example of spatial understanding, case 2.}
\end{center}

\clearpage
\noindent
\textbf{Robotics-Centric Understanding Case 1.}

\begin{center}
    \includegraphics[width=0.92\textwidth]{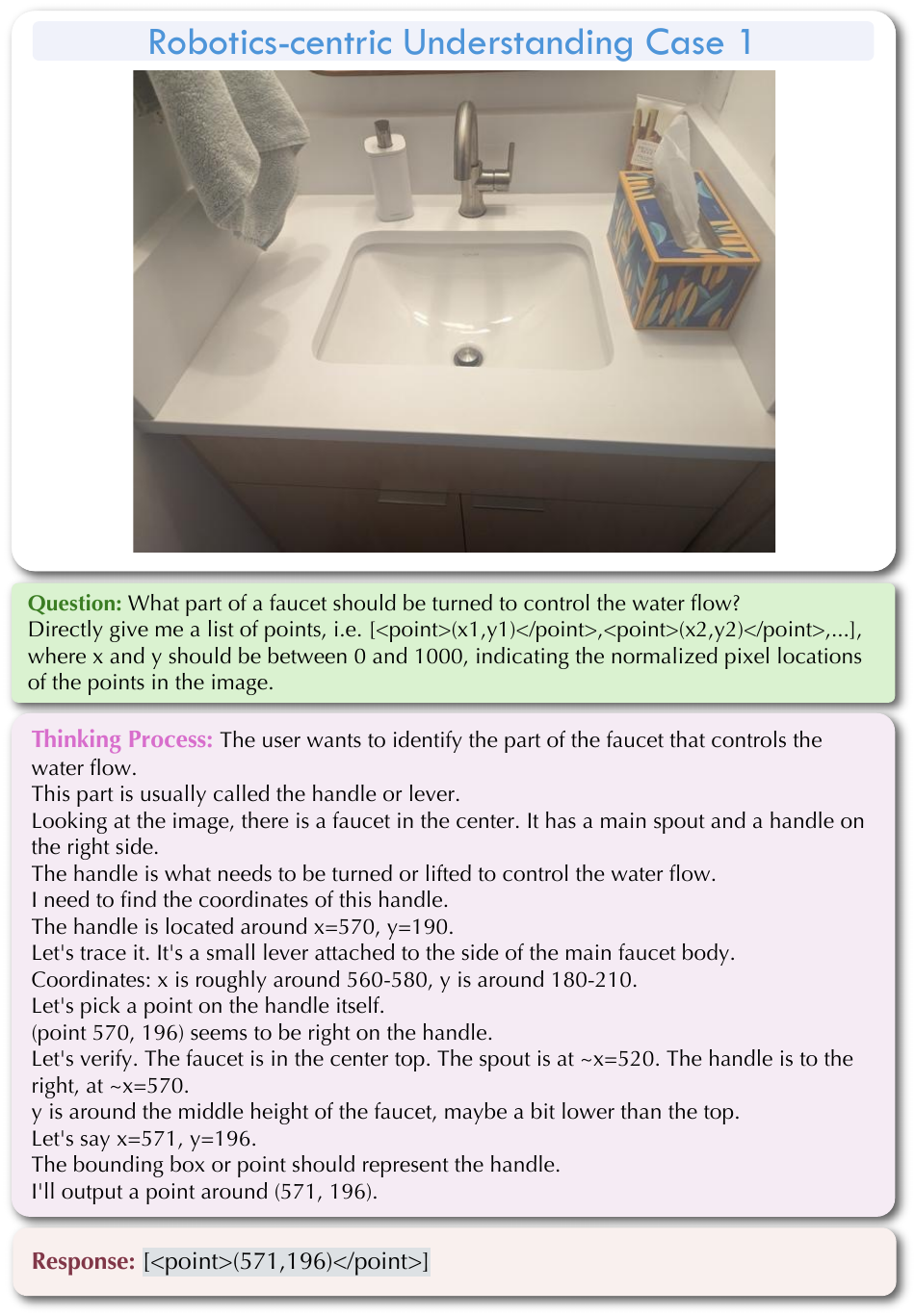}

    \vspace{0.5em}
    \refstepcounter{appendixfigure}
    \small Fig.\theappendixfigure: \textbf{Example of robotics-centric understanding, case 1.}
\end{center}

\clearpage
\noindent
\textbf{Robotics-Centric Understanding Case 2.}

\begin{center}
    \includegraphics[width=0.92\textwidth]{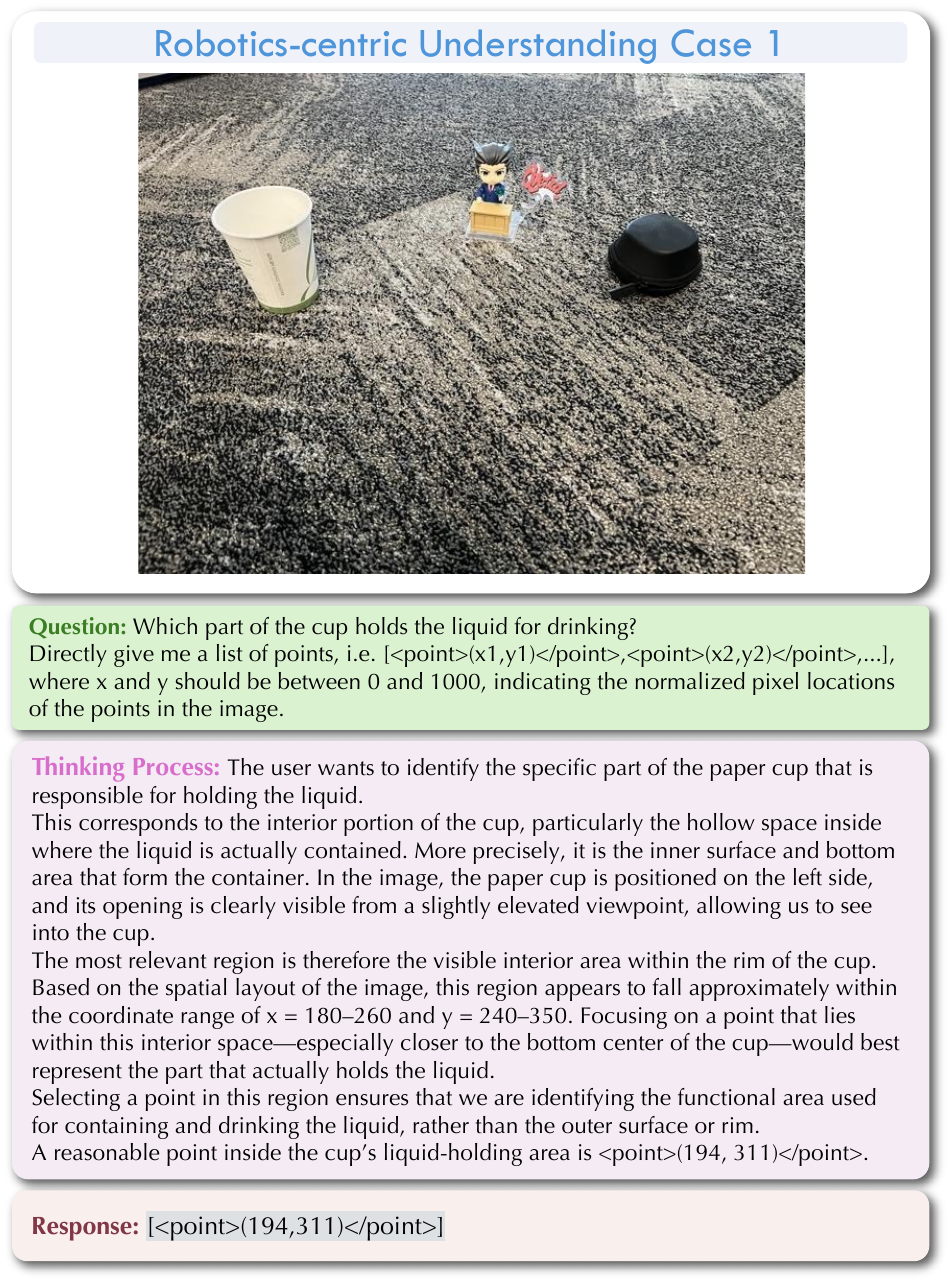}

    \vspace{0.5em}
    \refstepcounter{appendixfigure}
    \small Fig.\theappendixfigure: \textbf{Example of robotics-centric understanding, case 2.}
\end{center}

\clearpage
\noindent
\textbf{Interactive Understanding Case 1.}

\begin{center}
    \includegraphics[width=0.92\textwidth]{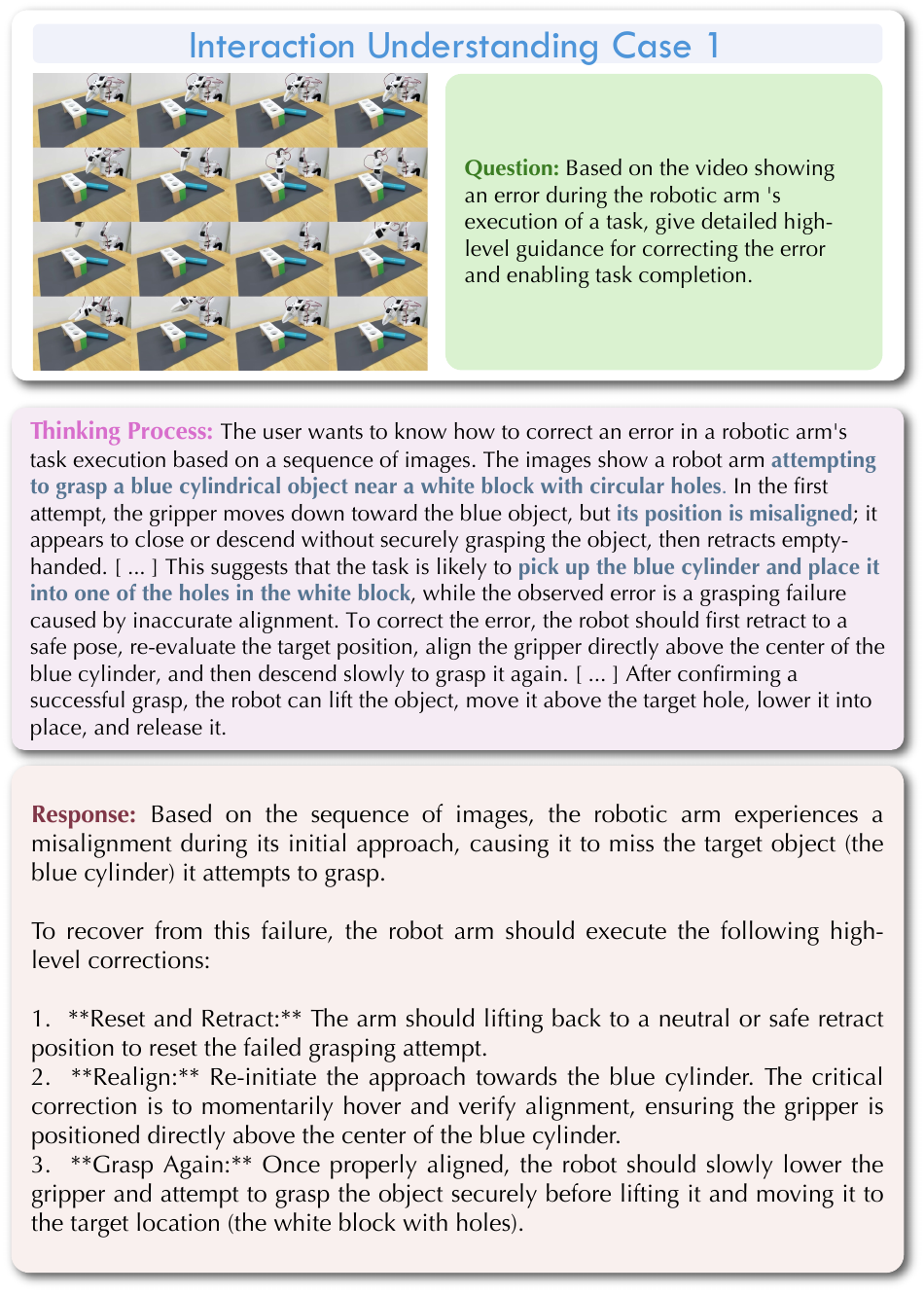}

    \vspace{0.5em}
    \refstepcounter{appendixfigure}
    \small Fig.\theappendixfigure: \textbf{Example of interactive understanding, case 1.}
\end{center}

\clearpage
\noindent
\textbf{Interactive Understanding Case 2.}

\begin{center}
    \includegraphics[width=0.92\textwidth]{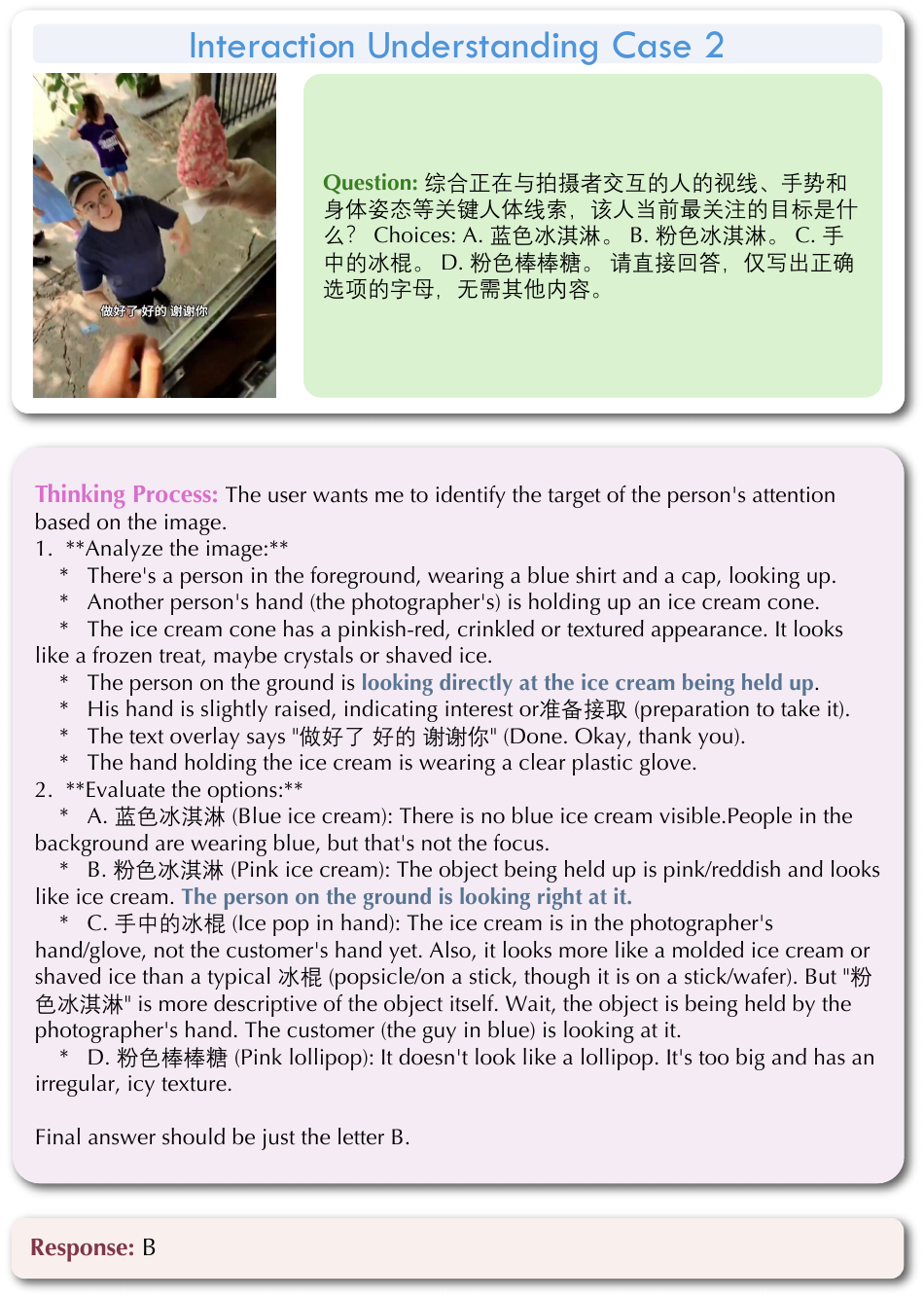}

    \vspace{0.5em}
    \refstepcounter{appendixfigure}
    \small Fig.\theappendixfigure: \textbf{Example of interactive understanding, case 2.}
\end{center}

\clearpage
\noindent
\textbf{Action Decision and Grounding Case 1.}

\begin{center}
    \includegraphics[width=0.92\textwidth]{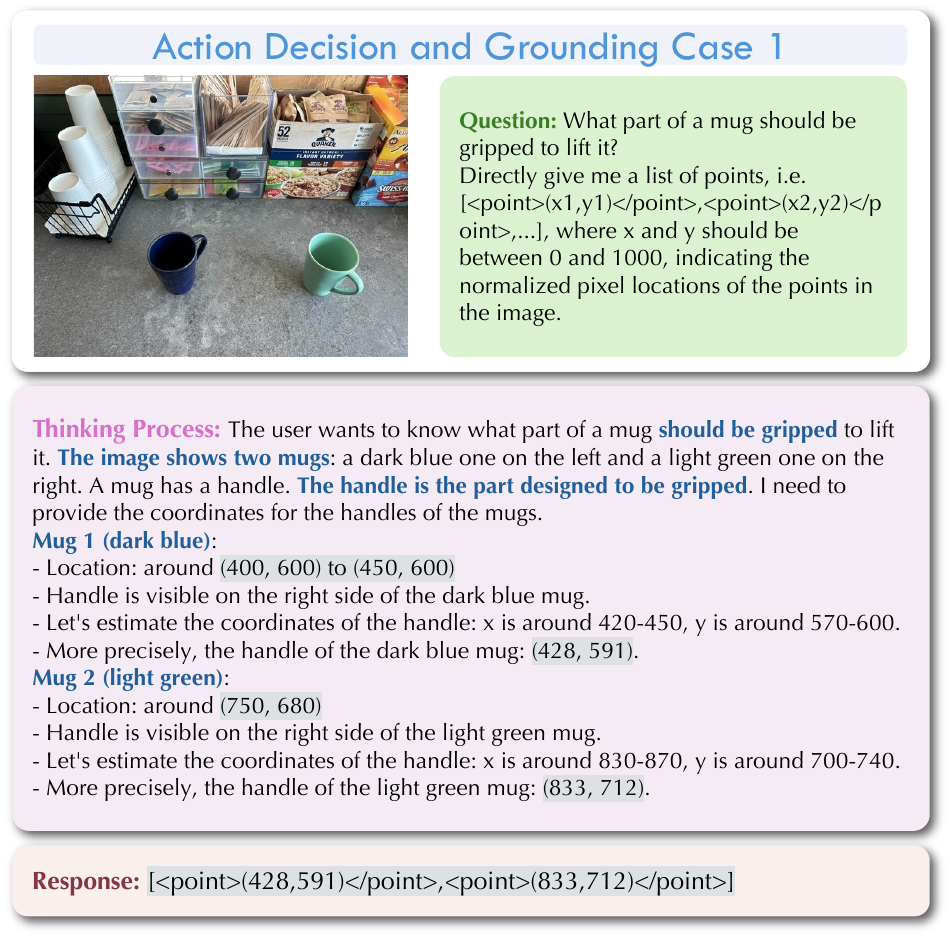}

    \vspace{0.5em}
    \refstepcounter{appendixfigure}
    \small Fig.\theappendixfigure: \textbf{Example of action decision and grounding, case 1.}
\end{center}

\clearpage
\noindent
\textbf{Action Decision and Grounding Case 2.}

\begin{center}
    \includegraphics[width=0.92\textwidth]{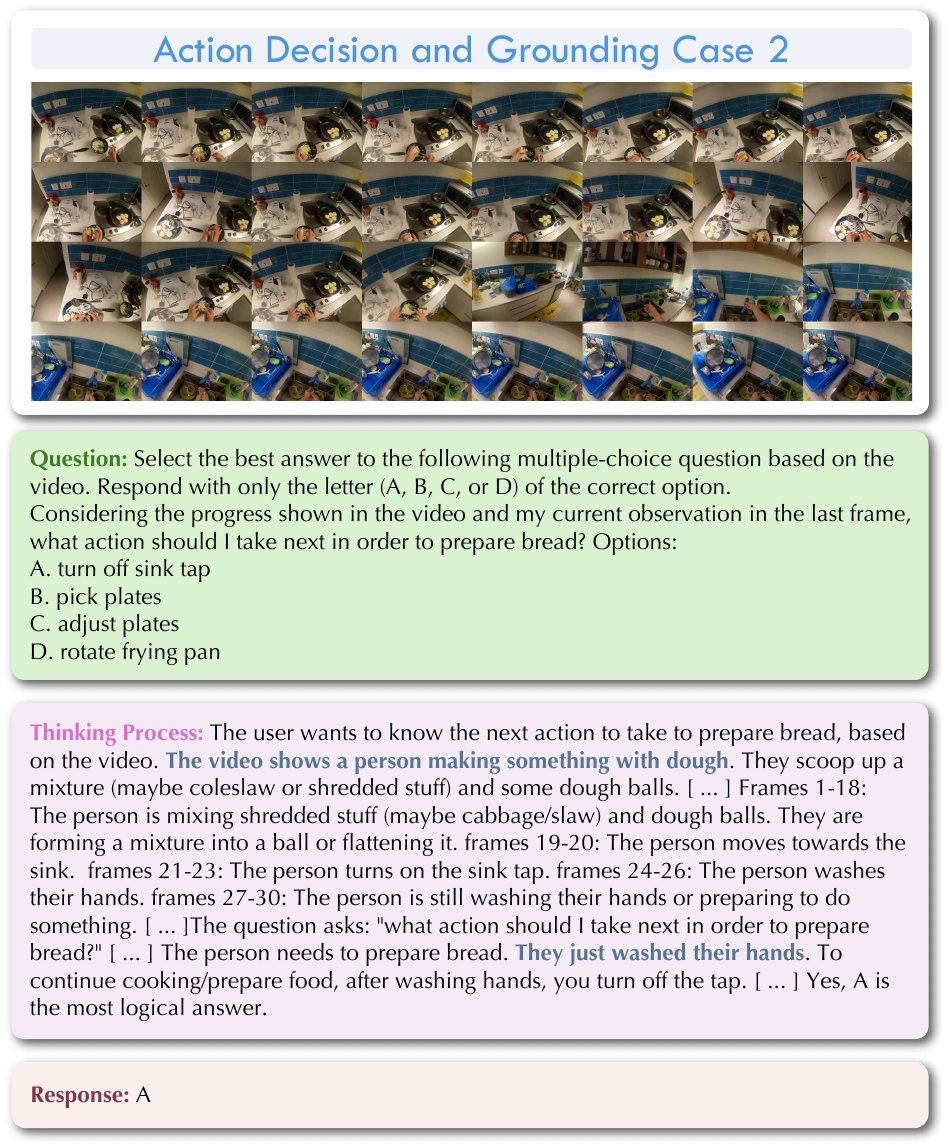}

    \vspace{0.5em}
    \refstepcounter{appendixfigure}
    \small Fig.\theappendixfigure: \textbf{Example of action decision and grounding, case 2.}
\end{center}

\clearpage
\noindent
\textbf{Physical Feasibility and Local Effects Case 1.}

\begin{center}
    \includegraphics[width=0.92\textwidth]{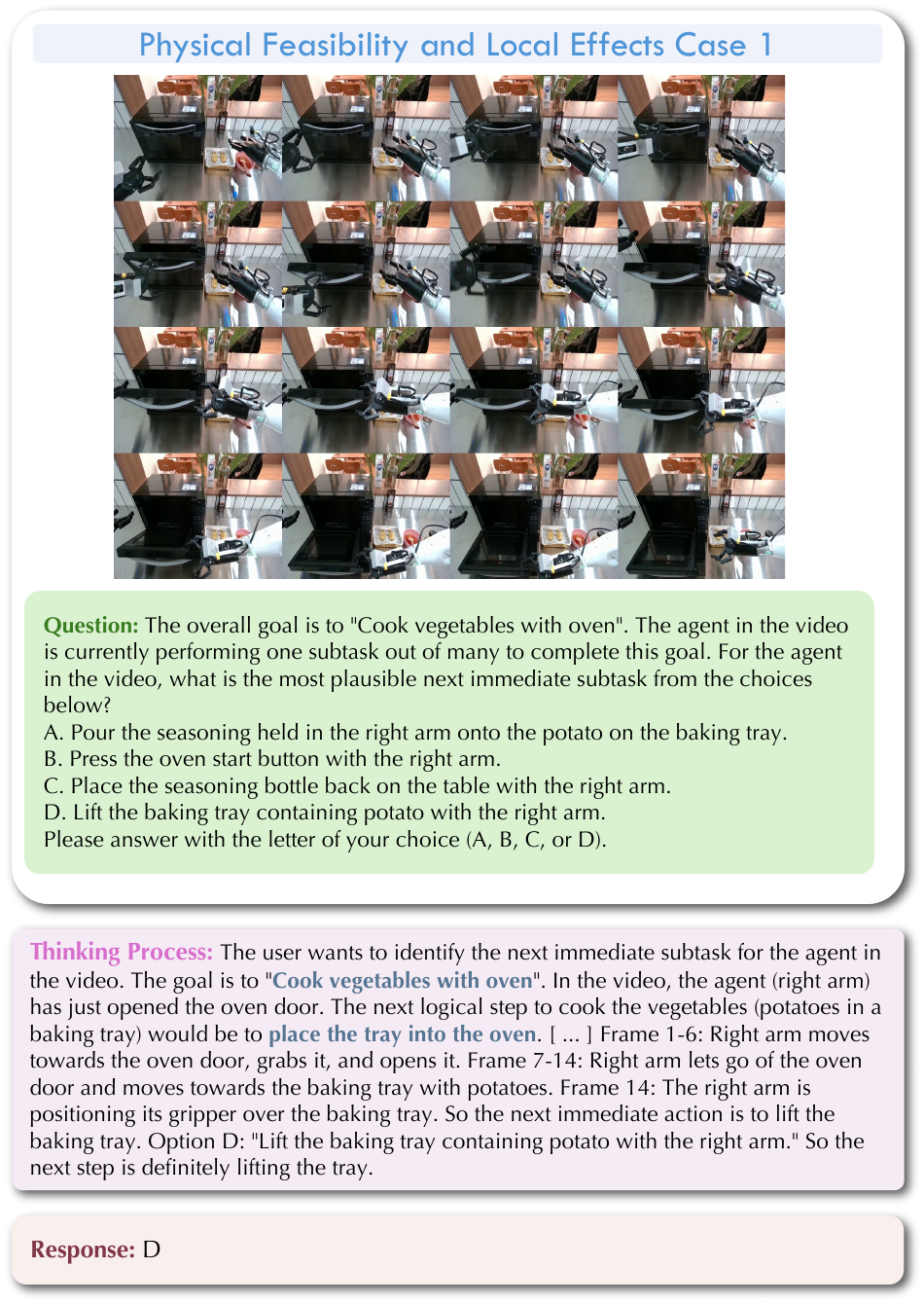}

    \vspace{0.5em}
    \refstepcounter{appendixfigure}
    \small Fig.\theappendixfigure: \textbf{Example of physical feasibility and local effects, case 1.}
\end{center}

\clearpage
\noindent
\textbf{Physical Feasibility and Local Effects Case 2.}

\begin{center}
    \includegraphics[width=0.92\textwidth]{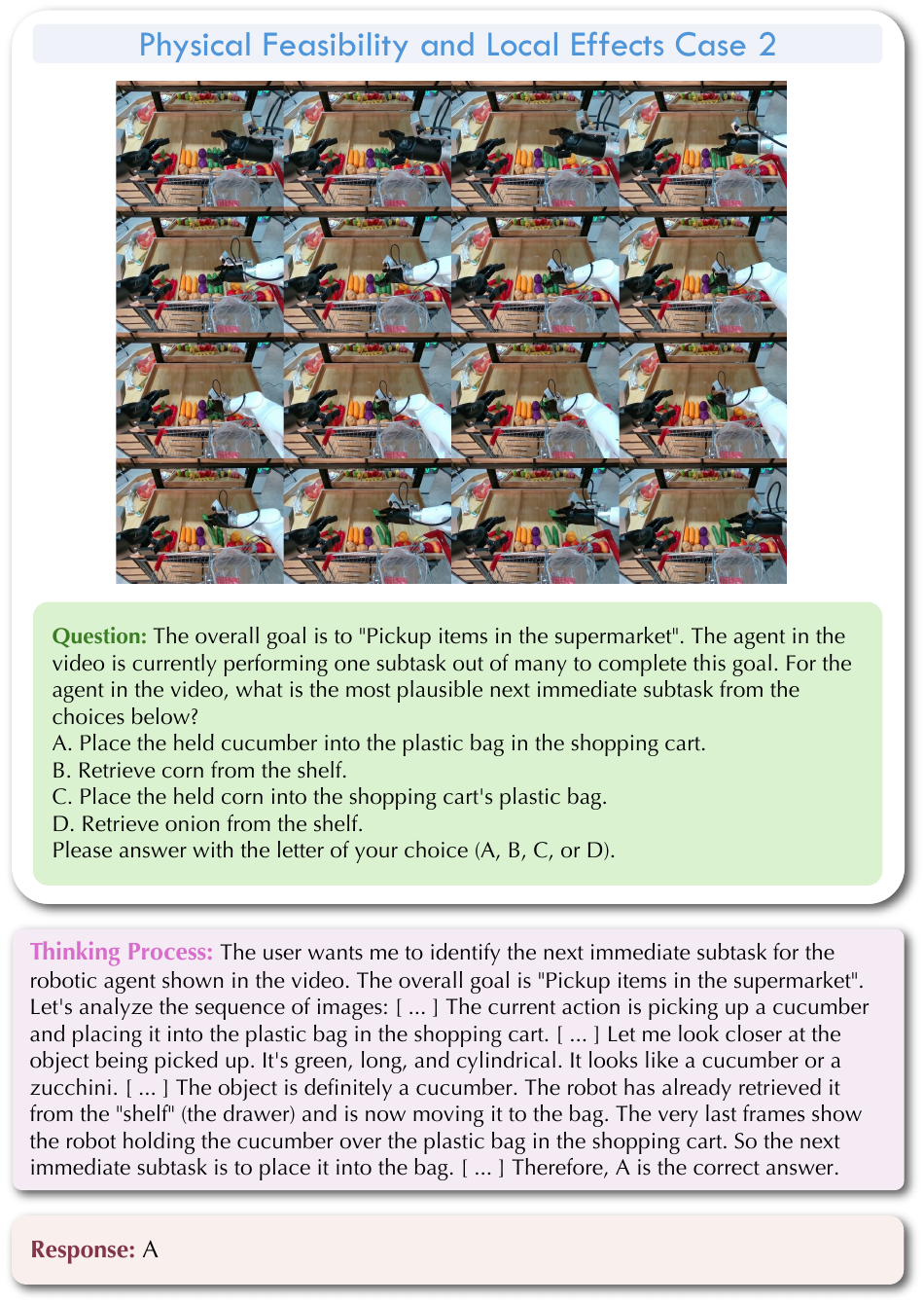}

    \vspace{0.5em}
    \refstepcounter{appendixfigure}
    \small Fig.\theappendixfigure: \textbf{Example of physical feasibility and local effects, case 2.}
\end{center}

\clearpage
\noindent
\textbf{Long-Horizon Planning and Composition Case 1.}

\begin{center}
    \includegraphics[width=0.92\textwidth]{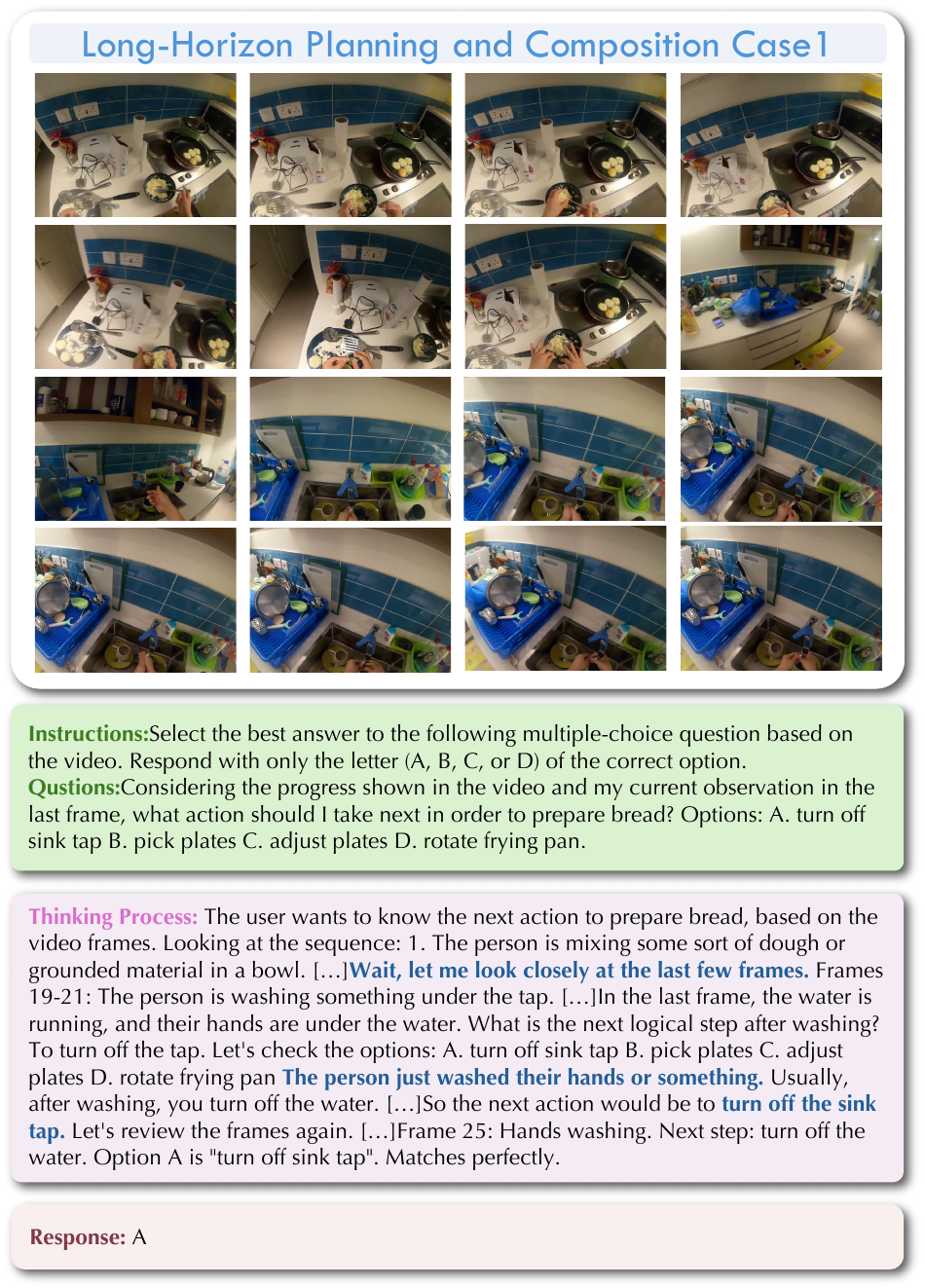}

    \vspace{0.5em}
    \refstepcounter{appendixfigure}
    \small Fig.\theappendixfigure: \textbf{Example of long-horizon planning and composition, case 1.}
\end{center}

\clearpage
\noindent
\textbf{Long-Horizon Planning and Composition Case 2.}

\begin{center}
    \includegraphics[width=0.92\textwidth]{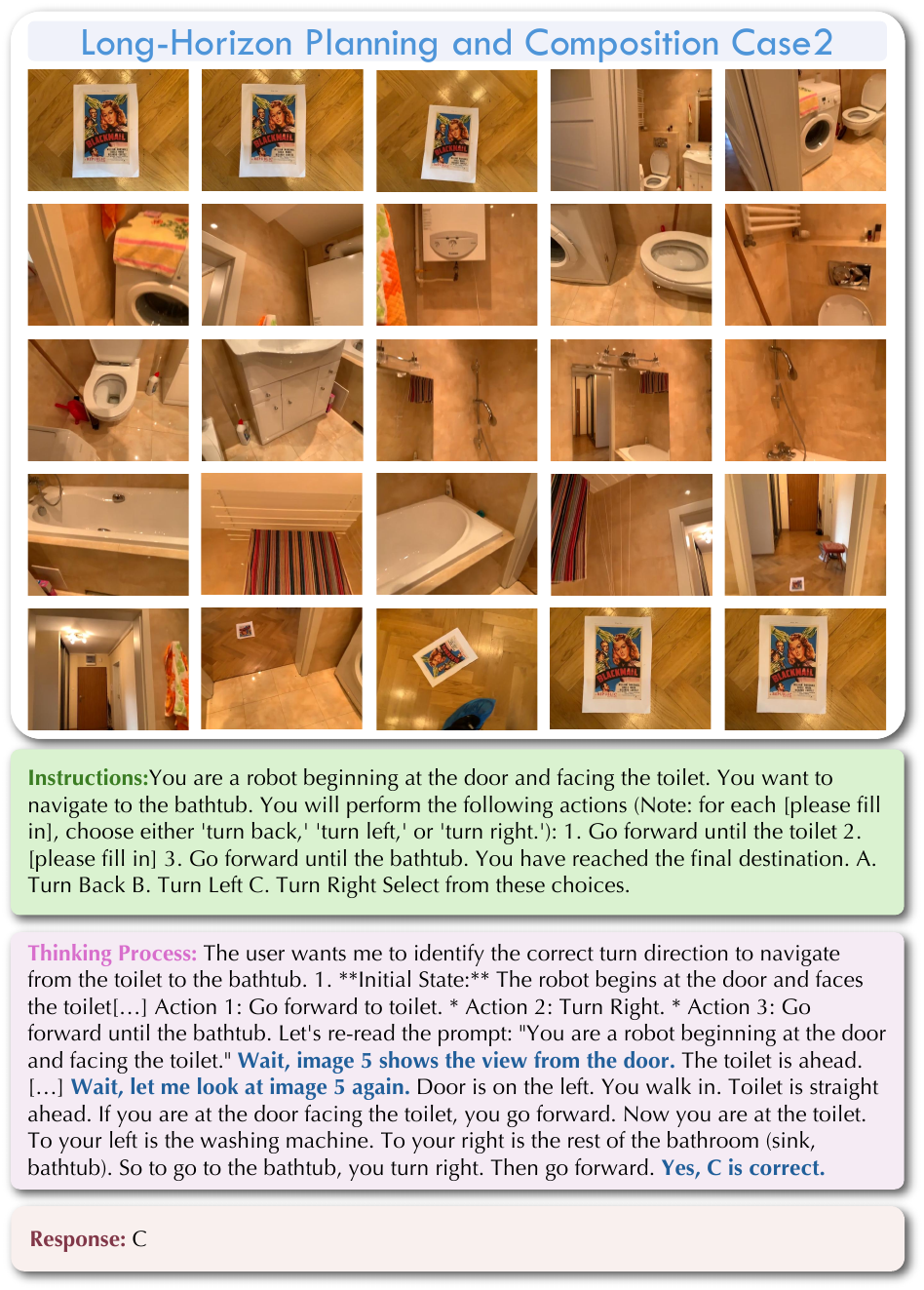}

    \vspace{0.5em}
    \refstepcounter{appendixfigure}
    \small Fig.\theappendixfigure: \textbf{Example of long-horizon planning and composition, case 2.}
\end{center}

\clearpage
\noindent
\textbf{Progress-Aware Sequential Reasoning Case 1.}

\begin{center}
    \includegraphics[width=0.92\textwidth]{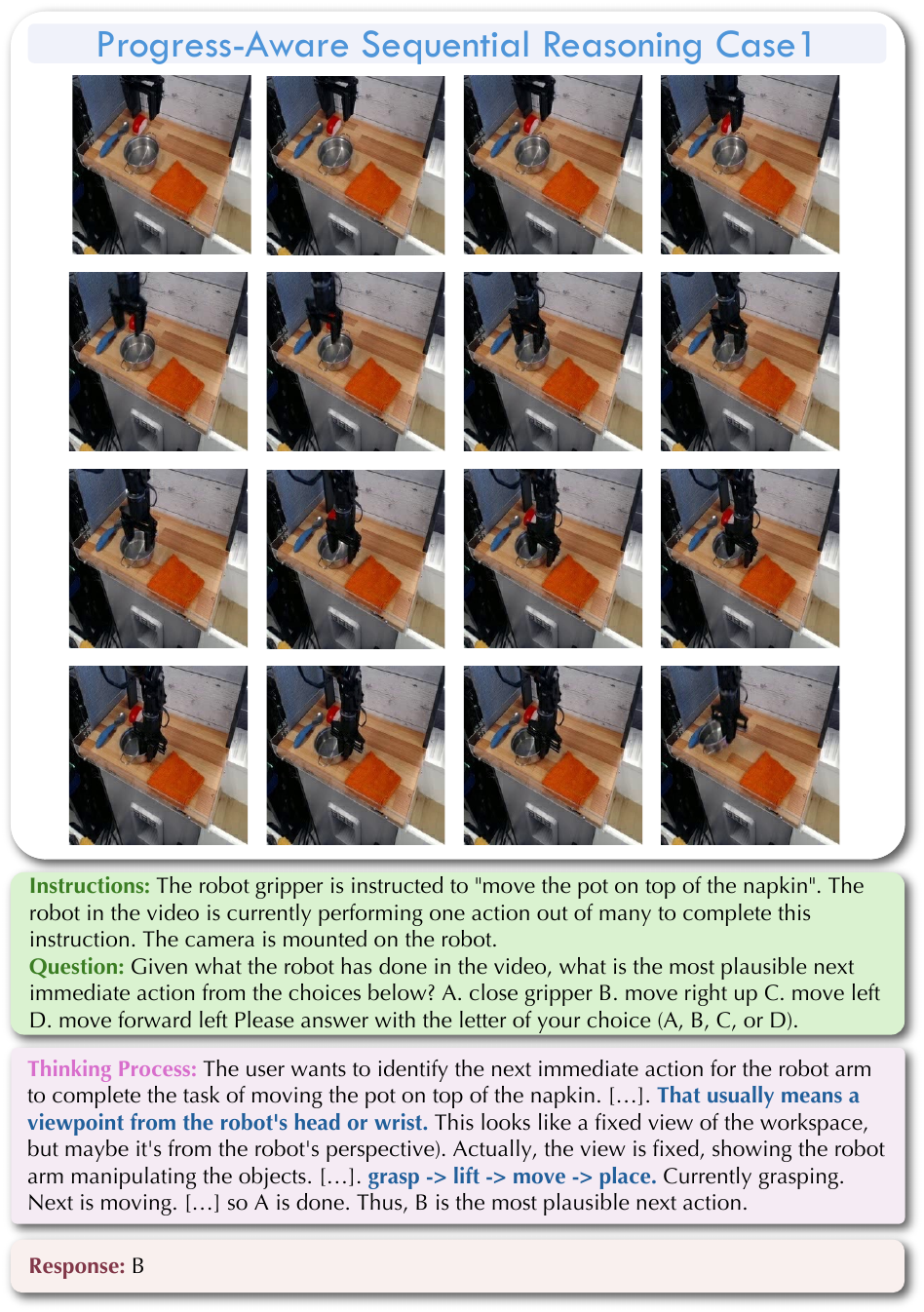}

    \vspace{0.5em}
    \refstepcounter{appendixfigure}
    \small Fig.\theappendixfigure: \textbf{Example of progress-aware sequential reasoning, case 1.}
\end{center}

\clearpage
\noindent
\textbf{Progress-Aware Sequential Reasoning Case 2.}

\begin{center}
    \includegraphics[width=0.92\textwidth]{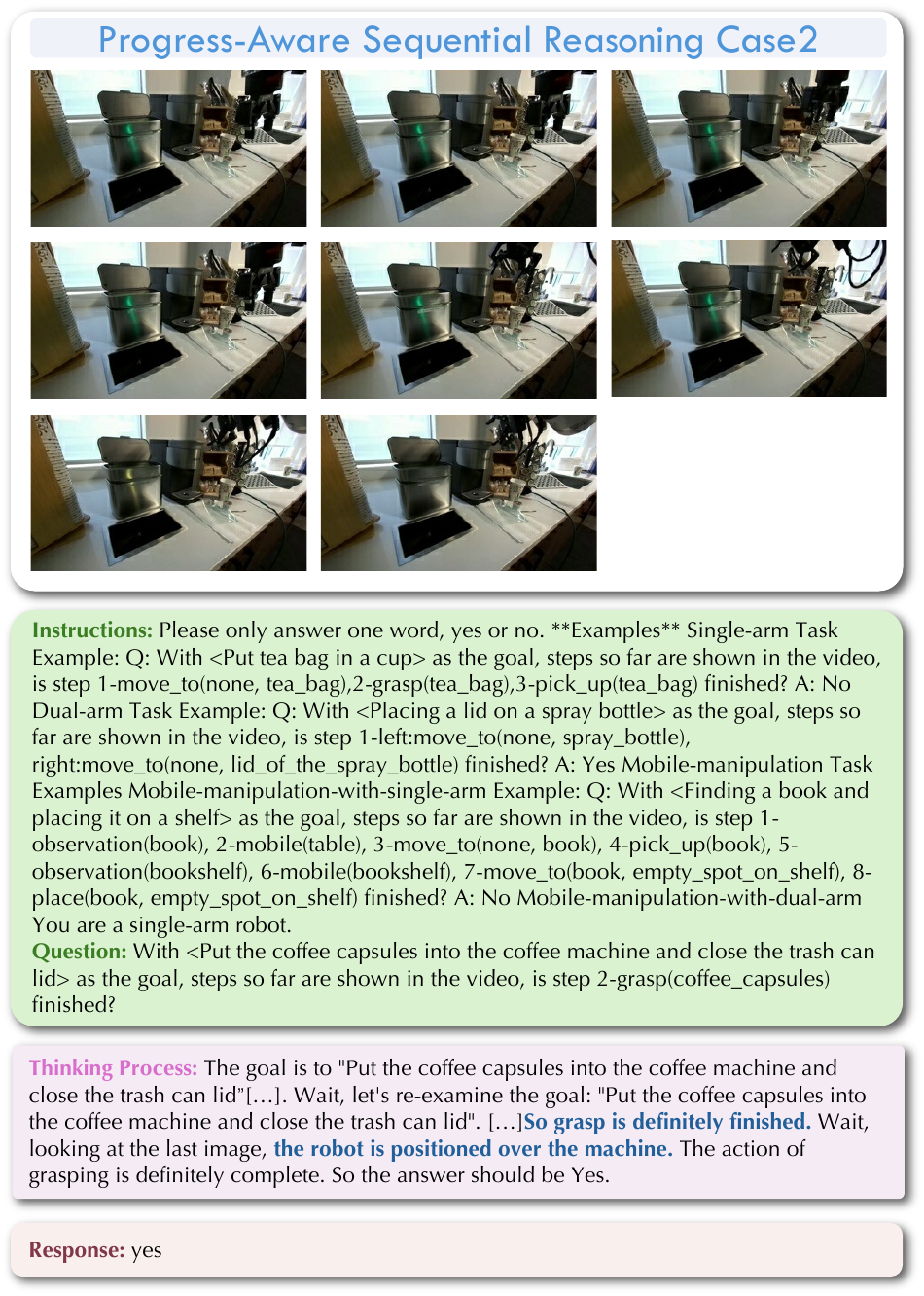}

    \vspace{0.5em}
    \refstepcounter{appendixfigure}
    \small Fig.\theappendixfigure: \textbf{Example of progress-aware sequential reasoning, case 2.}
\end{center}

\clearpage
\noindent
\textbf{Reflection, Repair, and Recovery Case 1.}

\begin{center}
    \includegraphics[width=0.92\textwidth]{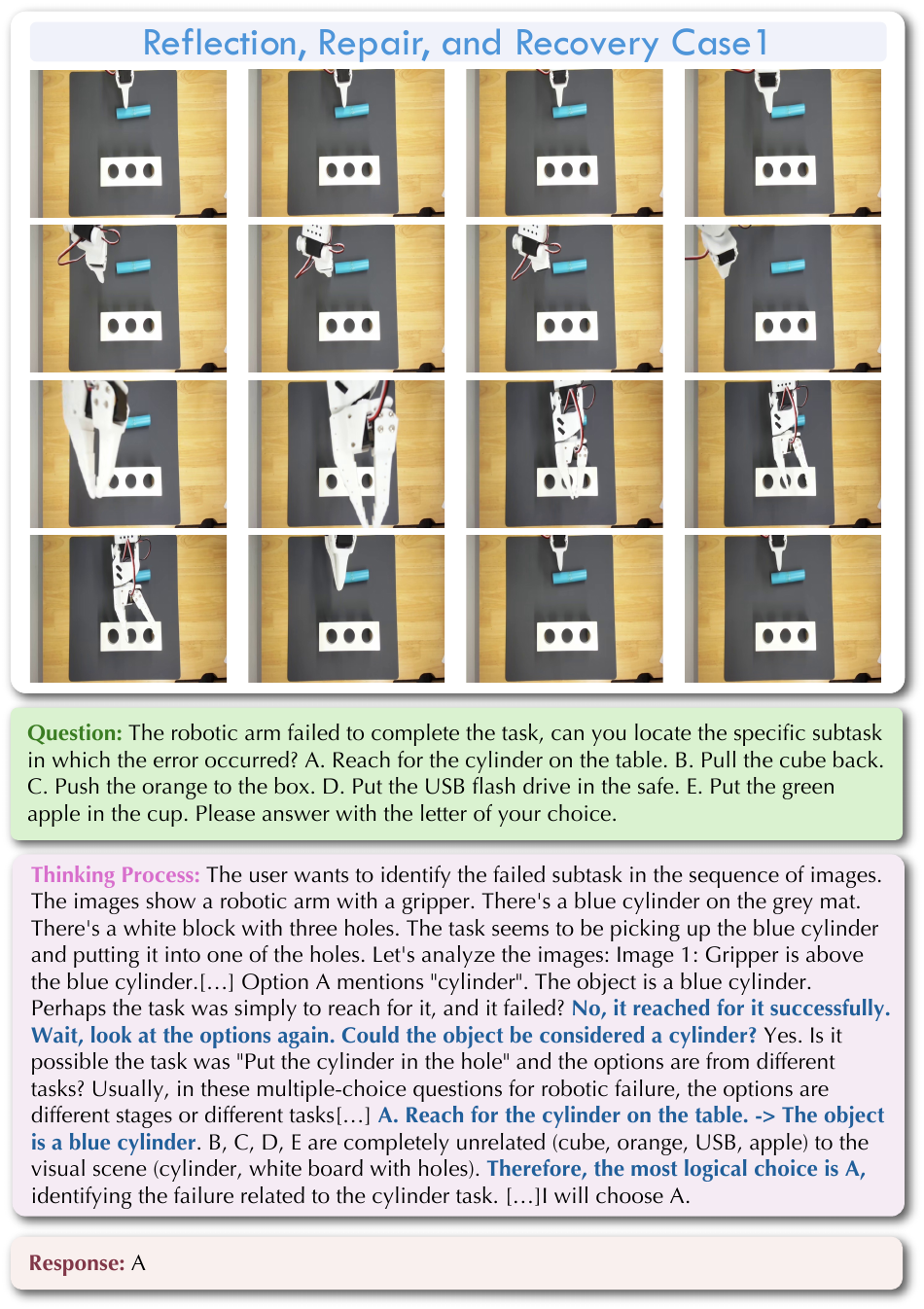}

    \vspace{0.5em}
    \refstepcounter{appendixfigure}
    \small Fig.\theappendixfigure: \textbf{Example of reflection, repair, and recovery, case 1.}
\end{center}

\clearpage
\noindent
\textbf{Reflection, Repair, and Recovery Case 2.}

\begin{center}
    \includegraphics[width=0.92\textwidth]{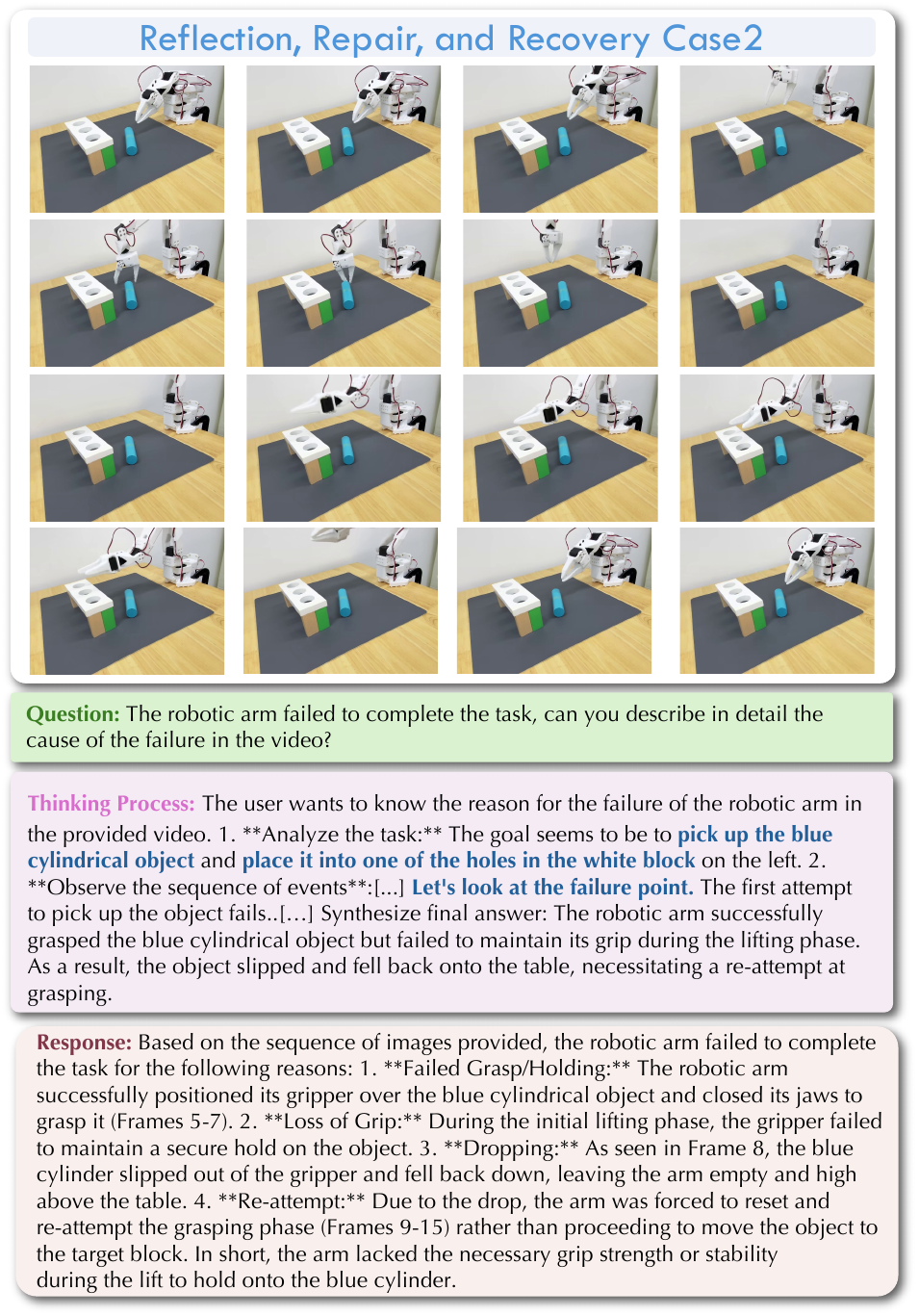}

    \vspace{0.5em}
    \refstepcounter{appendixfigure}
    \small Fig.\theappendixfigure: \textbf{Example of reflection, repair, and recovery, case 2.}
\end{center}

\clearpage
\noindent
\textbf{Vision-Language Navigation Case 1.}

\begin{center}
    \refstepcounter{appendixfigure}\label{fig:r2r-ce-qualitative}
    \includegraphics[width=0.92\textwidth]{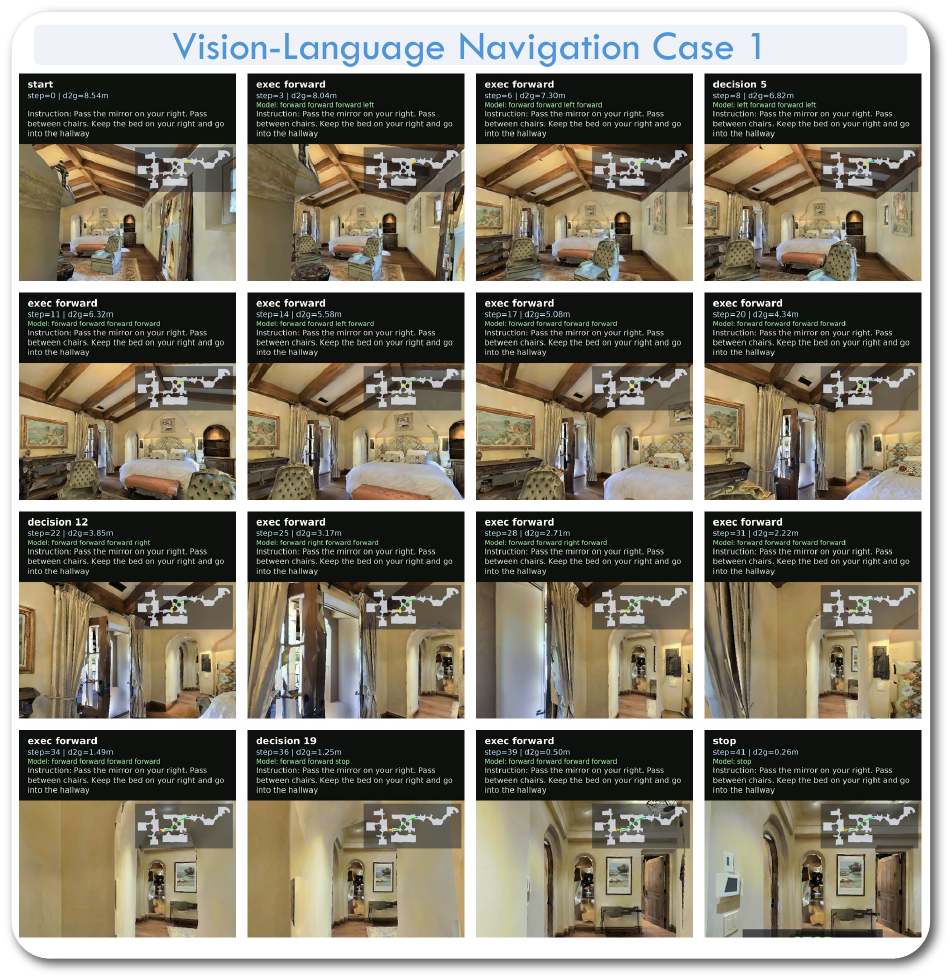}

    \vspace{0.5em}
    \small Fig.\theappendixfigure: \textbf{Successful R2R-CE \textit{val-unseen} episode.}
\end{center}

\clearpage
\noindent
\textbf{Object-Goal Navigation Case 1.}

\begin{center}
    \refstepcounter{appendixfigure}\label{fig:mp3d-objectnav-qualitative}
    \includegraphics[width=0.92\textwidth]{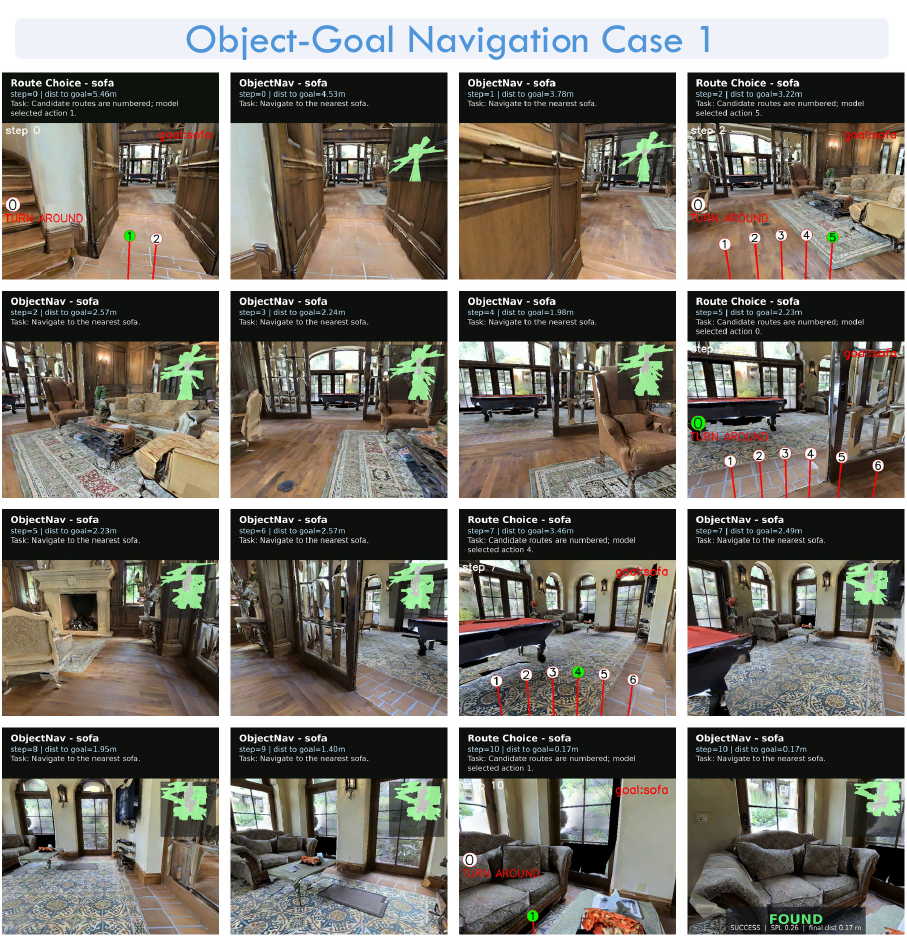}

    \vspace{0.5em}
    \small Fig.\theappendixfigure: \textbf{Successful zero-shot MP3D ObjectNav episode (target: sofa).}
\end{center}

\end{document}